\documentclass{article}

\usepackage{microtype}
\usepackage{graphicx}
\usepackage{subfigure}
\usepackage{booktabs} 

\usepackage{hyperref}

\usepackage[accepted]{icml2024}

\usepackage{amsmath}
\usepackage{amssymb}
\usepackage{mathtools}
\usepackage{amsthm}
\usepackage{soul}

\usepackage[capitalize,noabbrev]{cleveref}

\theoremstyle{plain}
\newtheorem{theorem}{Theorem}[section]

\newtheorem{lemma}[theorem]{Lemma}

\theoremstyle{definition}

\theoremstyle{remark}

\newcommand{\AK}[1]{{#1}}

\usepackage[textsize=tiny]{todonotes}

\icmltitlerunning{Error Feedback Can Accurately Compress Preconditioners}

\DeclareMathOperator{\Tr}{Tr}
\renewcommand{\algorithmiccomment}[1]{#1} 

\begin{document}

\twocolumn[

\icmltitle{Error Feedback Can Accurately Compress Preconditioners}



\icmlsetsymbol{equal}{*}

\begin{icmlauthorlist}
\icmlauthor{Ionut-Vlad Modoranu}{ista}
\icmlauthor{Aleksei Kalinov}{ista}
\icmlauthor{Eldar Kurtic}{ista}
\icmlauthor{Elias Frantar}{ista}
\icmlauthor{Dan Alistarh}{ista}
\end{icmlauthorlist}
\icmlaffiliation{ista}{Institute of Science and Technology Austria}
\icmlcorrespondingauthor{Ionut-Vlad Modoranu}{ionut-vlad.modoranu@ist.ac.at}

\icmlkeywords{Machine Learning, ICML}

\vskip 0.3in
]



\printAffiliationsAndNotice{}  

\begin{abstract}
Leveraging second-order information about the loss at the scale of deep networks is one of the main lines of approach for improving the performance of current optimizers for deep learning. 
Yet, existing approaches for accurate full-matrix preconditioning, such as Full-Matrix Adagrad (GGT) or Matrix-Free Approximate Curvature (M-FAC) suffer 
from massive storage costs when applied even to small-scale models, as they must store a sliding window of gradients, whose memory requirements are multiplicative in the model dimension. 
In this paper, we address this issue via a novel and efficient error-feedback technique that can be applied to compress preconditioners by up to two orders of magnitude in practice, without loss of convergence. 
Specifically, our approach compresses the gradient information via sparsification or low-rank compression \emph{before} it is fed into the preconditioner, feeding the compression error back into future iterations. 
Experiments on deep neural networks show that this approach can compress full-matrix preconditioners to up to 99\% sparsity without  accuracy loss, effectively removing the memory overhead of full-matrix preconditioners such as GGT and M-FAC. Our code is available on our GitHub repository \href{https://github.com/IST-DASLab/EFCP/}{https://github.com/IST-DASLab/EFCP/}.
\end{abstract}

\vspace{-1em}
\section{Introduction} \label{section:introduction}
The remarkable success of stochastic gradient descent (SGD)-based optimizers in deep learning motivated a long line of research for 
accelerated preconditioned variants that can still scale to the massive parameter and dataset sizes common in the area. Among these, optimizers based on adaptive regularization, such as Adagrad~\cite{Duchi2010AdaptiveSM} and Adam~\cite{kingma2014adam} are extremely well-established, with myriads of extensions, e.g.~\cite{agarwal2016secondorder, agarwal2019efficient, yao2020adahessian, gupta2018shampoo}. 

Yet, most work in this area restricts  preconditioning of the descent direction to \emph{a diagonal matrix}. 
By contrast, a promising---but considerably less developed---direction employs what we will call \emph{full-matrix preconditioning}, i.e. pre-multiplying the descent direction with a full matrix, whose structure is justified either via adaptive regularization~\cite{agarwal2019efficient}, or via approximations of natural gradient such as the Fisher approximation~\cite{1998-amari, 2021-MFAC}. 

Yet, existing implementations of full-matrix preconditioning are currently impractical given their massive memory costs. 
Interestingly, although they have different theoretical justifications, both full-matrix adaptive preconditioning methods such as GGT~\cite{agarwal2019efficient} and full-matrix natural-gradient methods such as M-FAC~\cite{2021-MFAC} run into the same key practical limitation: 
each maintains a ``sliding window'' of several past gradients, employed in estimating the preconditioner at each step. Even for small-scale models, maintaining the gradient history results in unmanageable total cost of $\Theta(md)$ memory, where $d$ is the model dimension, usually in the millions, and $m$ is the size of the gradient history, usually of size 100-1000. 
For example, a standard recipe for fine-tuning the medium-scale BERT-base model using the M-FAC preconditioner with standard settings requires more than 450GB of GPU RAM,  unavailable on any single GPU. 
Thus, while showing promising results in terms of accuracy, full-matrix preconditioning is currently impractical at scale due to memory constraints.

\paragraph{Contribution.} 
In this paper, we present a new algorithmic approach for reducing the massive memory cost of full-matrix preconditioning at the scale of DNNs, called Error Feedback for Compressed Preconditioning (EFCP), 
which applies to both adaptive (Adagrad/GGT~\cite{agarwal2019efficient}) and natural-gradient (M-FAC~\cite{2021-MFAC}) full-matrix preconditioners. 
In a nutshell, EFCP works by compressing the gradient window either via high sparsity or extremely low rank. 
In practice, we are able to compress the gradient history by one to two orders of magnitude \emph{without impacting accuracy}, effectively removing the memory limitations of full-matrix preconditioning. 

Our results are enabled by two technical innovations. 
On the conceptual side, the surprising observation behind our work is that the sizable error in the preconditioner induced by gradient sparsification or low-rank projection can in fact be handled via an error feedback mechanism applied to the gradients \emph{before they are stored} into the gradient window. 
For both variants, we provide efficient compression-aware data structures which can leverage compression for memory savings, and can apply to both M-FAC and GGT. We also provide partial theoretical justification for the good convergence properties of our algorithm. 

On the practical side, our main contribution is an efficient GPU implementation for EFCP in the case of high sparsity, which realizes our theoretical gains in practice. Specifically, this is enabled by a novel GPU implementation of a \emph{dynamic sparse ring-buffer} data structure, in which each element is a sparse gradient tensor, and which enables efficient sparse multiplication against new gradients, as needed for preconditioner computations, as well as fast gradient insertions and removals, as required by the sliding window semantics. 

We validate our implementation experimentally on standard vision (ResNet/ImageNet) and language modeling (BERT/GLUE) tasks. 
Results show that EFCP enables us to significantly reduce the memory overhead of full-matrix preconditioning, to the point where it is in the same range as the memory required by standard optimizers such as SGD with Momentum or Adam. 

We show this extensively by testing the EFCP implementation in the context of the M-FAC optimizer, compressing gradients to $\sim 99\%$ sparsity (a variant we call Sparse MFAC), as well as via smaller-scale experiments for low-rank compression and the GGT full-matrix Adagrad optimizer. 
For example, for the standard ResNet-18 model~\cite{ResNet} \textit{trained from scratch} on ImageNet~\cite{russakovsky2015imagenet}, the Sparse MFAC variant with full-matrix preconditioning uses a maximum of 22.5 GB of GPU RAM, compared to 21.3 GB used by SGD with momentum with the same parameters. At the same time, Sparse MFAC outperforms the well-tuned SGD baseline  by approximately 1\% Top-1 test accuracy, recovering the accuracy of Dense M-FAC in the same setup (which uses more than 56GB GPU RAM in this setting). 
We provide additional validation for Sparse M-FAC in the context of training ViT~\cite{2020-dosovitskiy} and BERT~\cite{2019-devlin} models as well, and perform ablations w.r.t. sparsity and model size. 

In summary, for the first time, EFCP enables \AK{researchers to experiment} with full-matrix preconditioner methods within the memory capacity of a single GPU in reasonable time, with negligible accuracy loss relative to full preconditioning. 

\section{Background and Related Work} \label{section:background-and-related-work}

\subsection{General Setting}
We consider a standard classification setting, in which we are given a training dataset $\mathcal{D}=\{(x_i, y_i)_{i=1}^N\}$ containing $N$ i.i.d. samples $x_i$ and their associated labels $y_i$. We suppose input pairs $(x,y)$ are drawn from a true, unknown distribution $p_{true} \triangleq p_t$ having the density $p_t(x,y) = p_t(x)p_t(y|x)$ and our model $\theta \in \mathbb{R}^d$ parameterizes a probabilistic distribution $p_\theta(y|x) \triangleq p(y|f(x, \theta))$ with density $p_\theta(x,y) = p_t(x) p_\theta(y|x)$.
Our aim is to minimize the ``difference'' between the true conditional distribution $p_t(y|x)$ and the modelled conditional distribution $p_\theta(y|x)$, 
measured in some appropriate metric. If we pick this to be the KL divergence between the two distributions, we obtain the following loss function $KL(p_t(x,y) || p_\theta(x,y)) = \int p_t(x,y) \log \frac{p_t(x,y)}{ p_\theta(x,y)}dx dy = \mathbb{E}_{p_t(x)}[KL(p_t(y|x)||p_\theta(y|x))]$.

In practice, we only have a finite number of samples $x \sim p_t(x)$ and no information about the density function. In this case, we are using the empirical training distribution $\hat{p}_t$ instead of $p_t$, which is given by the samples in $\mathcal{D}$ and thus we can define the loss as $L(\theta) \triangleq KL(p_t(x,y) || p_\theta(x,y)) = \mathbb{E}_{\hat{p}_t(x)} [KL(p_t(y|x)||p_\theta(y|x))] \approx -\frac{1}{N} \sum_{i=1}^N \log p_\theta(y_i|x_i),$ which is the standard objective function minimized for the maximum likelihood estimation. 
The loss $L:\mathbb{R}^d \rightarrow \mathbb{R}$ is assumed to be continuous and twice differentiable.

In practice, the loss $L(\theta)$ defined above is often minimized using variants of Stochastic Gradient Descent (SGD) $\theta_{t+1} \gets \theta_t - \eta_t g_t$, where $g_t \triangleq \nabla_\theta L(\theta_t)$ and $\eta_t \in \mathbb{R}$ is the learning rate.
However, theoretical results suggest that convergence and end-results can be improved by \emph{preconditioning} this gradient with a matrix $C$ which incorporates information about the geometry of the loss landscape, leading to the parameter update rule $\theta_{t+1} \gets \theta_t - \eta_t C_t^{-1} g_t$.

\paragraph{Natural Gradient.} 

\AK{Instead of using the Euclidean distance to adapt to the information geometry, Natural Gradient~\cite{1998-amari} relies on KL-divergence to measure the discrepancy between the true distribution and the modelled distribution.
In this context,} the preconditioner $C$ takes the form of the Fisher Information Matrix (FIM), defined as 
$F(\theta) \triangleq \frac{1}{N} \sum_{i=1}^N \mathbb{E}_{p_\theta(y|x_i)} [\nabla_\theta \log p_\theta(y|x_i) \nabla_\theta \log p_\theta(y|x_i)^T].$ 
One can show that the FIM is the expected Hessian with respect to the log-likelihood. 
Since the exact FIM is difficult to compute due to the expectation term in its definition, different efficient approximations of it exist in the literature. 
For neural networks, K-FAC~\cite{grosse2016kroneckerfactored, martens2018kroneckerfactored} provides a diagonal and tri-diagonal approximation of FIM by sampling a limited number of $y$ values from $p_\theta(y|x_i)$ in order to compute the expectation.

Another popular approximation to the FIM is the Empirical Fisher, defined as $\hat{F}(\theta) \triangleq \frac{1}{N} \sum_{i=1}^N \nabla_\theta \log p_\theta(y_i|x_i) \nabla_\theta \log p_\theta(y_i|x_i)^T$. Instead of sample $y$'s from model's predictive distribution, it uses the associated label $y_i$. It is important to note that the empirical Fisher is a good approximation of the FIM if $p_t(y|x) \approx p_\theta(y|x)$, e.g. when the model has good accuracy. 
The M-FAC method~\cite{2021-MFAC} (Section~\ref{sec:detailed-methods}) provides an efficient implementation for this approach. 

\paragraph{Adaptive Regularization.} 
A different justification for obtaining good preconditioners is via \emph{adaptive regularization}, arising from online convex optimization. 
Broadly, the idea is to adapt the learning rate corresponding to each parameter in order to minimize regret, 
defined as the cumulative difference between the optimizer's performance and the best decision in hindsight, 
by taking into account the history of gradients observed up to a given point. 
By and large, the community has focused on diagonal preconditioners, which allow computationally-efficient and extremely popular algorithms, such as Adam~\cite{kingma2014adam} and AdaGrad~\cite{Duchi2010AdaptiveSM}. 
The GGT result~\cite{agarwal2019efficient} (Section~\ref{sec:detailed-methods}) showed that results can be improved by taking ``second-order'' weight correlations into account, leading to a full-matrix, $d \times d$ preconditioner, and provided an efficient approximation of this preconditioner.

\subsection{Efficient Full-Matrix Preconditioners}
\label{sec:detailed-methods}

\paragraph{Overview.} 
We now provide a detailed description of the M-FAC full-matrix preconditioner~\cite{2021-MFAC}, which will serve as the main illustration of our preconditioner compression method in the paper body. 
A fully-worked-out example for the GGT preconditioner~\cite{agarwal2019efficient} is also provided in the Appendix. 
A common point of both methods is that they store a history of the past $m$ gradients in the matrix 

\begin{equation}
    G = [g_{t-(m-1)}^T, g_{t-(m-2)}^T, ..., g_{t-1}^T, g_t^T] \in \mathbb{R}^{m \times d}, 
\end{equation} which can be represented as a ring buffer, updated at each step by replacing the oldest gradient $g_{t-(m-1)}$ with the most recently-generated one $g_t$. 
\AK{M-FAC and GGT then use the ring buffer $G$ differently to approximate the preconditioner.}
To preserve performance, the buffer $G$ must be stored in the GPU memory. 
For best accuracy, both methods require storing a large window of $m$ gradients, where $m$ is usually between $100$ and $1000$, each of which are of size $d$ (the model dimension). Thus, the memory cost of each method becomes intractable in large-scale settings. Next, we briefly describe the algorithmic implementation of M-FAC. 

\paragraph{The M-FAC Optimizer.} \label{preconditioner:M-FAC} M-FAC provides an efficient estimator for Inverse-Hessian-Vector-Products (IHVPs) for the special case where the Hessian is approximated via the Empirical Fisher approximation $\hat{F} = \frac{1}{m} \sum_{i=1}^m g_i g_i^T$, where the gradients $g_t \triangleq \nabla_\theta L(\theta_t)$ are stored in the matrix $G$, defined above. 
The first idea is to use the classic Woodbury-Sherman-Morrison (WSM) formula to integrate new gradient information as rank-1 updates into the current inverse. This leads to the following recursion to compute the IVHPs with an arbitrary vector $x \in \mathbb{R}^d$ as 
\begin{eqnarray*}
\hat{F}_{t+1}^{-1}x = \left(\hat{F}_t + \frac{1}{m}g_t g_t^T\right)^{-1}x  \\ \stackrel{WSM}{=\joinrel=\joinrel=} \hat{F}_t^{-1}x - \frac{\hat{F}_t^{-1} g_t (\hat{F}_t^{-1} g_t)^T}{m + g_t^T \hat{F}_t^{-1} g_t}x  \\ = \frac{1}{\lambda}x - \sum_{k=1}^t \frac{\hat{F}_k^{-1} g_k (\hat{F}_k^{-1} g_k^T)}{m + g_k^T \hat{F}_k^{-1} g_k}x,
\end{eqnarray*}
where $F_0 = \lambda I_d$ and $\lambda > 0$ is a damping parameter. 

The key technical idea behind M-FAC is the authors' non-trivial observation that IHVPs can be expressed as a linear combination between the past gradients in the sliding window, where the coefficients are efficiently computed based only on scalar products $g_i^T g_j$ and $g_i^T x$, leading to the following expression for the IHVPs as 
\begin{equation} \label{formula:mfac-linear-combination}
    \hat{F}_m^{-1}x = \frac{1}{\lambda}x - \sum_{k=1}^m c_k g_k.
\end{equation}
This allows M-FAC to efficiently add or remove gradients from the sliding window, done by computing the corresponding coefficients $c_k$, and then making the resulting update into the formula for $\hat{F}_m^{-1}x$. 

A key aspect regarding M-FAC is the interaction with regularization. Recent work~\cite{ThreeMechanisms} explains that weight decay and $L_2$-regularization coincide for SGD, but are different for preconditioned methods. Simply inputting $g_t = \nabla_\theta L(\theta) + \gamma \theta$ into the M-FAC preconditioner invalidates the empirical Fisher definition due to the additional term $\gamma \theta$. Based on this, and differring from M-FAC~\cite{2021-MFAC}, we will decouple the regularization such that the empirical Fisher definition holds and use the update $\theta_{t+1} = (1-\gamma\eta_t) \theta_t - \eta_t \hat{F}_t^{-1} \nabla_\theta L(\theta_t)$, where $\gamma > 0$ is the weight decay parameter.

\paragraph{Data Structures and Operations.} 
Next, we briefly summarize the main operations in the original M-FAC optimizer (denoted next by Dense M-FAC). We detail the challenges that appear when the internal buffer $G$ is sparse, as well as our GPU-efficient solution, in the further sections. 
First, notice that the main operations in M-FAC are the Scalar Products (SP) and the Linear Combination of Gradients (LCG), which are needed to express the IHVPs as a linear combination.

\paragraph{Scalar Products (SP).} This first operation corresponds to the pairwise scalar products of rows in $G \in \mathbb{R}^{m \times d}$ stored in an intermediary matrix $S = GG^T \in \mathbb{R}^{m \times m}$, where $S_{ij} = g_i^Tg_j$. When the buffer $G$ is updated (e.g. the oldest gradient is replaced by a new gradient $g \in \mathbb{R}^{d}$ at row $i$), the $i^{th}$ row and $i^{th}$ column of matrix $S$ are updated as $S_{i,:} = S_{:,i} = G \cdot g \in \mathbb{R}^m$. This operation can be easily performed on GPU using frameworks such as Pytorch.

\paragraph{Linear Combination of Gradients (LCG).} The second main operation of M-FAC is the linear combination of gradients, having the coefficients vector $c \in \mathbb{R}^m$. For a \emph{dense} buffer $G$, the LCG is easily computed by multiplying each component $c_i$ with the $i^{th}$ row of $G$ and then computing the sum of all rows in $G$. This operation can also be easily performed in PyTorch on GPU.


\subsection{Related Methods} 

In the Appendix, we provide a full discussion of GGT~\cite{agarwal2019efficient}, as well as its compressed version, which can be seen as a full-matrix version of AdaGrad optimizer. 
More generally, obtaining efficient approximations of second-order (curvature) information in deep learning is an active area. 
The \emph{diagonal} approximation of the Hessian~\cite{krishnan2017neumann, kingma2014adam} is very popular; 
however, it is known to provide lower quality relative to both block-wise methods or K-FAC~\cite{2019-wang, 2020-singh, 2021-MFAC}. 

K-FAC provides a block diagonal approximation of the FIM, 
and allows efficient computation of the inverse~\cite{2016-ba, Osawa_2019,2019-zheng, 2019-wang,laurent2018an}; 
however, it is known that its prerequisites do not always hold~\cite{2015-martens}. 
\emph{Hessian-free} optimization~\cite{martens_free} forgoes the explicit computation of Hessians in favor of computing an IHVP, 
but may require several ``inner'' iterations to converge at each step. 
AdaHessian~\cite{yao2020adahessian} provided an alternative approximation of the inverse Hessian diagonal, using Hutchinson's randomized algorithm for estimating the diagonal, which requires smoothing heuristics, and has at least 2x theoretical iteration cost vs SGD, which in practice is usually around 3x. 
Shampoo~\cite{gupta2018shampoo} provides an efficient heuristic approximation of GGT, reducing memory cost by maintaining preconditioner estimates per model tensor, using an efficient iterative method to compute preconditioner inverses.

\paragraph{The Error Feedback Mechanism.}
The error feedback mechanism has been studied extensively in the context of gradient compression for standard SGD, 
in the absence of complex preconditioning. 
The first analyses of sparsified SGD with error feedback were provided by~\cite{stich2018sparsified, 2018-alistarh} under additional assumptions, while~\cite{2019-karimireddy, stich2019error, nadiradze2020elastic} provided additions to the analysis. More recently, a very complex argument by~\citet{li2022distributed} analyzed convergence of diagonal Adagrad with error feedback. We show that their analysis extends to a diagonal version of our algorithm. 

\section{Method} \label{section:method}

\begin{algorithm}[!ht]
    \caption{EFCP: Error Feedback for Accurate Compressed Full-Matrix Preconditioning}\label{algorithm:generic-efcp}
    \begin{algorithmic}[1]
        \STATE Parameters: $T=$steps count; $m=$ gradient count;
        \STATE $\xi_0 = 0_d\in \mathbb{R}^d$  \COMMENT{$\triangleright$ initialize error feedback}
        \FOR{each step $t \in \{1, 2, ... T\}$}
            \STATE $g_t \gets \nabla_\theta L(\theta_t)$  \COMMENT{$\triangleright$ get gradient}
            \STATE \textcolor{blue}{$a_t \gets \xi_{t-1} + g_t$} \COMMENT{$\triangleright$ error feedback}
            \STATE \textcolor{blue}{$c_t \gets \textsc{Compress}(a_t)$} \COMMENT{$\triangleright$ compress the accumulator}
            \STATE \textcolor{blue}{$\xi_t \gets a_t - c_t$} \COMMENT{$\triangleright$ update the error}
            \STATE $u_t \gets \mathcal{A}(\textcolor{blue}{c_t})$ \COMMENT{$\triangleright$ update $G$ and precondition using $\mathcal{A}$}
            \STATE $\theta_{t} \gets \theta_{t-1} -\eta_t u_t$ \COMMENT{$\triangleright$ update model parameters}
        \ENDFOR
    \end{algorithmic}
\end{algorithm}

\paragraph{Overview.} Our general approach for compressing the gradient history required by full-matrix preconditioners is described in Algorithm~\ref{algorithm:generic-efcp}. 
Conceptually, we assume a standard gradient-based optimization setup with an oracle $g_t$ at each step $t$, augmented by a preconditioning algorithm $\mathcal{A}$, based on a gradient sliding window $G$, 
whose large memory cost we want to reduce. 
We propose to do this by feeding \emph{compressed} gradients into the preconditioner estimate maintained by algorithm $\mathcal{A}$. Yet, doing so directly by projecting the gradients onto a sparse or low-rank support diverges.

Instead, we apply \emph{error feedback}~\cite{2014-seide, 2018-alistarh, 2019-karimireddy} to gradients \emph{before they are fed into the preconditioner data structure}, using an error accumulator vector $\xi_t \in \mathbb{R}^d$, initially set to zero. At each  step, we obtain the gradient $g_t$ of the model parameters w.r.t. the loss, after which we feed back the error $\xi_{t-1}$ into the current gradient $g_t$, obtaining the \emph{accumulator} $a_t$.  
Next, the accumulator $a_t$ is compressed, using e.g. sparsity or low-rank, to obtain a compressed representation $c_t$. 
The compression error $\xi_t$ is updated to the difference between the accumulator $a_t$ and its compressed version $c_t$. 
Finally, the compressed gradient $c_t$ is fed directly to the preconditioning algorithm $\mathcal{A}$. 

Notice that, other than the three lines corresponding to error feedback (\textcolor{blue}{blue} in the algorithm), the optimization loop is the same as the uncompressed one. However, we emphasize that the internal buffer $G$ of the preconditioned optimizer will store a sequence of $m$  representations $[c_{t-(m-1)}, c_{t-(m-2)}, ..., c_{t-1}, c_t]$, each of which is highly-compressed, leading to memory savings.

While this blueprint appears simple, two key questions remain: 
(1) \emph{how can we leverage the compressed gradient window for actual memory and computational speedups on complex real-world preconditioners?} and 
(2) \emph{why does this approach preserve convergence?} 
We address them below. 


\subsection{Compressing Preconditioners via Sparsity} \label{subsection:efcp-top-k}

We now expand on accumulator compression via sparsity, specifically by using Top-$k$ sparsification and error feedback. Specifically, the  preconditioner data structure will only store $k \ll d$ entries from each accumulated gradient, 
where $k$ is around $1\%$ of the full dimension $d$. As we will show experimentally, this will essentially remove the memory overhead of full-matrix preconditioning. 

We detail our approach for the Sparse M-FAC algorithm; the Sparse GGT variant is slightly simpler and is described in Appendix~\ref{section:appendix-sparse-ggt} due to space limitations.
For Sparse M-FAC, we describe the changes required by sparsity for the buffer $G$ and the Scalar Product (SP) and Linear Combination of Gradients (LCG) operations. 
A key concern will be re-interpreting the mathematical formulation in order to be able to leverage fast CUDA kernels, which is notoriously challenging in the context of sparsity. 

\paragraph{Sparsity format for $G$.} 
To store the gradient history $G$, we need a sparse matrix-like data structure for storing gradients that fulfills the following two properties at the same time: i) it allows easy replacement of any row (corresponding to adding and removing gradients) and ii) it allows computing the scalar product (SP) and linear combination of gradients (LCG) operations efficiently. To the best of our knowledge, no such data structure is known in the literature. 

Specifically, for Sparse M-FAC, we want to input sparse accumulators $c_t$ of size $k \ll d$ into the buffer $G$. Instead of storing $G \in \mathbb{R}^{m \times d}$ densely, we will store $G=(\mathcal{I}, \mathcal{V})$,  where $\mathcal{I} \in \mathbb{N}^{m \times k}$ will store the indices returned by Top-$k$ and $\mathcal{V} \in \mathbb{R}^{m \times k}$ will store the corresponding values. When the $j^{th}$ row of $G$ is updated with the new indices-values pair $(i_j, v_j)$, we replace the $j^{th}$ row in $\mathcal{I}$ and $\mathcal{V}$ with $i_j$ and $v_j$, respectively. Breaking $G$ into two matrices of indices and values means that $G$ loses the matrix structure in the dense sense and the SP and LCG operations must be redefined because we cannot use the basic PyTorch operations for matrix multiplications anymore. Below we describe our solution for these two operations that involve CUDA specific kernels to efficiently perform the computations on a GPU for large models, but first we would like to provide details about the sparsity pattern which we can leverage to obtain more efficient kernels for our particular problem.

\paragraph{Sparsity Pattern.} One observation is that, due to the algorithm structure, each row in $G=(\mathcal{I}, \mathcal{V})$ has the same sparsity. To leverage this, at each step we compress the accumulator $a_t$ using Top-$k$ applied in blocks of (large)  fixed size $B$. 
The block separation distributes compressed values more uniformly, speeding up the LCG operation.
We express $k$ as a percentage of total number of weights $d$ and then convert it to a particular integer value that represents the density of values used in the Top-$k$ call. For example, in most of our experiments we will have $k=1\%$, meaning that for each block of size $B$ we will have {exactly $\lceil B / 100 \rceil$} non-zero values. This way we know that for any block of size $B$ we have the same number of weights in a specific interval. We provide more details about choosing $B$ in the section where we describe the LCG operation.


\paragraph{Format for $\mathcal{V}$.} Starting with the CUDA capability 8.0, the Nvidia GPUs have bfloat16 support, which is a half precision data type optimized for Deep Learning. We found that our optimizer still converges when we use bfloat16 instead of float32. This is particularly useful when automatic mixed precision (AMP) is activated because it speeds up the overall training and has a lower memory footprint (both for the model's state and for the optimizer's state).

\paragraph{Sparse Scalar Products.} This operation is commonly known as sparse-matrix vector multiplication (SPMV) and several implementations already exist for a few sparsity representations (e.g. CSR, COO), but unfortunately they cannot be directly used for our particular choice for the matrix $G=(\mathcal{I}, \mathcal{V})$, each of size $m \times k$. 
We implement an SPMV kernel that leverages the properties of our buffer. 
We use $m$ thread blocks, each containing 1024 threads. One thread block processes one row and has a shared memory array of 1024 floats to store the dot product computed on a subset of the row. We perform memory coalesced reads for the indices and values from the global memory. Lastly we perform parallel reduction over the shared memory buffer to compute the sum (the final value for the dot product) in logarithmic complexity. We show our implementation of the SP operation in the Appendix~\ref{section:appendix-cuda-kernels}.

\paragraph{Sparse LCG.} 
\AK{To make computation of linear combination of gradients efficient, we heavily rely on the limited block sizes $B$ with the fixed number of non-zero values in each block.
When Top-$k$ compression  is applied blockwise, each selected index indicates a relative position inside the block and not an absolute position in the large weight matrix. The limited range of possible index values, combined with the freedom to choose the block size, allows us to reuse these indices directly as positions in the shared memory. 
}
\AK{We} accumulate the linear combination of gradients across a slice of size $m \times B$ from $G=(\mathcal{I}, \mathcal{V})$ in shared memory and transfer the aggregated result in the output global memory at the correct offsets. Similar to the SP case, we also perform memory coalesced reads for the indices and values from the global memory only once for each component. We show our implementation of the LCG operation in the Appendix~\ref{section:appendix-cuda-kernels}.

\paragraph{Thread Blocks for LCG.} 
Choosing the right number of thread blocks is critical for the performance of the Sparse LCG, since it needs to balance memory transfer and compute load. 
To address this, we provide an automatic procedure for choosing this parameter, based on the GPU capabilities. 
Specifically, given a number of threads and maximum available GPU shared memory size, we determine the total number of thread blocks to be used to allow for full utilization, avoiding waiting in the queue at each SM.

\paragraph{Efficiency Improvements.} 
\AK{The described} technical data structure optimizations are critical for good performance. Relative to a naive Pytorch implementation of the same approach, which stores the gradients as sparse but performs the operations in dense format, the above approach is 1.5-2x faster and uses 20-30\% less memory. 

\paragraph{Memory Savings.}
Compressing the buffer using sparsity yields $\tilde{O}(mk)$ space complexity, where $k$ is the target density of the gradients after Top-$k$ compression, ignoring logarithmic factors due to indexing. In practice, we obtain stable convergence for $k = d / 100$, which translates into practical space savings of approximately $20$x (relative to GGT) and $30$x (relative to M-FAC). We provide exact numbers in the experimental section.

\subsection{Low-Rank Compression of Preconditioners}

\AK{We also implement} the $\textsc{Compress}$ step in Algorithm~\ref{algorithm:generic-efcp} via a low-rank approximation of the gradient, \AK{computed} using an efficient variant of power iteration~\cite{vogelsPowerSGDPracticalLowRank2019}. 
\AK{Similar to Top-$k$ compression, we reformulate algebraic operations of M-FAC to leverage low-rank matrices for memory and computational savings. } We provide implementation details in Appendix~\ref{app:low-rank} and \AK{additionally explain the relation between matrix decomposition rank $\rho$ and compression rate $k$ of Top-$k$ in Appendix~\ref{appendix:rank-topk-connection}. }

\subsection{Theoretical Justification}

Next, we turn our attention to theoretical justifications for our method.
These are detailed in Appendix~\ref{appendix:theory}, and summarized here. 
First, in the case where the preconditioning is only performed over the \emph{diagonal}, corresponding for instance to the AdaGrad optimizer~\cite{Duchi2010AdaptiveSM} but with error feedback, we can show that our approach recovers a convergence rate of $O( 1 / \sqrt T + \sigma^2 / \sqrt T + d / T )$ relative to the Adagrad baseline where gradients are uncompressed, by showing that it satisfies the preconditions of the analysis of~\citet{li2022distributed} for diagonal adaptive methods with gradient compression.

In the much more challenging full-matrix case, we provide an argument showing that our approach can indeed come close to the strong $O( \sqrt{d / T})$  average regret bound of full-matrix Adagrad~\cite{Duchi2010AdaptiveSM}, over $T$ rounds and for dimension $d$, under some additional ``smoothness'' assumptions on the rate of change of the Hessian approximation. 

In brief, our result says that, if 1) the Hessian approximation $H_t$ is stable from one iteration $t$ to another w.r.t. the error feedback, in the sense that $H_{t}H_{t-1}^{-1}\xi_{t} \simeq \xi_t$ for all $t \geq 1$, where $\xi_{t}$ denotes the error buffer; and 2) if the misalignment between stochastic gradients and the preconditioning direction is bounded over time, then Sparse M-FAC obtains $O( \sqrt{d / T} \cdot \sqrt{d / k} )$ average regret over $T$ rounds. 

Specifically, the additional regret due to sparse preconditioning is proportional to $\sqrt{d / k}$,  the square root of the fraction of the gradient $\ell_2$ norm preserved by the Top-$k$ truncation operation. While $\Theta ( \sqrt{d / k} )$ is a worst-case bound, in practice it is usually lower due to the fact that gradient values tend to be normally-distributed. 
Overall, this argument provides partial justification for our method, although further work is needed to fully understand the remarkable practical convergence of Sparse M-FAC, which we illustrate next.


\section{Experiments} \label{section:experiments}

We now validate our results experimentally. In the main text, we focus on comparing Sparse M-FAC with other optimizers, as it proved to be the most practical variant of EFCP. We show results for low-rank M-FAC and Sparse GGT in the Appendix. 

In the following, we validate Sparse M-FAC (S-MFAC) experimentally as an optimizer for training in the context of standard vision and language modelling tasks. 
Specifically, we integrate our S-MFAC optimizer in ASDL~\cite{ASDL}, a benchmarking library for second-order optimizers, but also examine larger-scale tasks such as 
ImageNet training of ResNet-18 
and compare S-MFAC with AdamW and Dense M-FAC (D-MFAC) on BERT models on GLUE tasks. We report training loss, test accuracy and total running times and top memory usage for the entire training process. 

\paragraph{Notations.} We use the notation $E$ for number of epochs, $\eta$ for learning rate, $\gamma$ for weight decay, $B$ for batch size, $m$ for the number of gradients (sliding window size), $k=1\%$ \AK{for} the gradient density. We use the prefixes \textbf{D} for \textbf{D}ense and \textbf{S} for \textbf{S}parse in the context of adaptive optimizers M-FAC. 
For the gradient window size $m$, we use the standard settings for D-MFAC~\cite{2021-MFAC}, which is  $m=1000$ unless otherwise specified.


\paragraph{The ASDL Benchmark.} In the first experiment, we integrate S-MFAC into the ASDL benchmarking framework for second-order optimizers~\cite{ASDL}, for the task of fine-tuning a ViT-T model pre-trained on ImageNet. We report the test accuracy for the run that has largest validation accuracy. Table~\ref{table:asdl-results} shows the test accuracy of our S-MFAC optimizer compared to the other optimizers reported in the paper. (We had to remove P-SGD from the comparison due to errors raised by the implementation during experiments.) We observe that S-MFAC is competitive with other preconditioned gradient methods, outperforming SGD, AdamW, K-FAC and SENG, and matching the highest accuracy obtained on this task (98\%) with affordable time and memory overheads relative to SGD. 

\begin{table}[!h]
\begin{center}
\caption{\label{table:asdl-results} Test accuracy of S-MFAC for ResNet-18 and ViT-Tiny on CIFAR-10. The accuracies for the other optimizers are obtained from ASDL~\cite{ASDL}. We also report relative time and memory overheads w.r.t. SGD.}
\begin{tabular}{c|cccc}
    \toprule
    Optimizer  & ViT-T & Time & Memory\\
    \midrule
    SGD        & 97.8 & $1\times$ & $1\times$\\
    AdamW      & 97.9 & $1\times$ & $1.01\times$\\
    K-FAC(1mc) & 97.4 & $1.12\times$ & $1.13\times$\\
    SENG       & 97.7 & $35\times$ & $7.31\times$\\
    Shampoo    & 98.0 & $1.6\times$ & $1.07\times$\\
    \midrule
    S-MFAC     & 98.0 & $1.63\times$ & $1.15\times$\\
    \bottomrule
\end{tabular}
\end{center}
\end{table}

\paragraph{ImageNet/ResNet-18.} Next, we move to a more complex vision task, by training ResNet-18 on ImageNet~\cite{ImageNet} \textit{from scratch} using a highly-tuned version of SGD (with momentum) from the FFCV~\cite{FFCV} library, as well as S/D-MFAC. All hyper-parameters are provided in the Appendix~\ref{appendix:hyper-parameters}. Figure \ref{figure:imagenet-resnet18} shows the validation accuracy curves for this experiment, while Figure~\ref{table:imagenet-resnet18} shows the best achieved training loss and validation accuracy for each method, including a detailed sparsity sweep below $1\%$ density for S-MFAC. The loss curves show similar trends, and are provided in the Appendix. 

We begin by noting that SGD obtains training loss \AK{with value of} $2.237$ and accuracy \AK{of} $69.12\%$. D-MFAC reaches the loss \AK{value of} $2.164$ and significantly higher Top-1 validation accuracy, $70.03\%$. Despite slightly higher training loss than D-MFAC ($2.185$), S-MFAC has comparable $70.01\%$ validation accuracy, which is strong for this model and training recipe. Thus, even for this challenging task, compressing the preconditioner essentially recovers the accuracy of the dense M-FAC optimizer. 
When we examine the behavior of S-MFAC as we decrease gradient density, we observe that, remarkably, S-MFAC with $k=0.17\%$ (only $\approx 20k$ entries!) still achieves comparable accuracy to SGD, and that accuracy decreases gradually. However, the resulting preconditioned gradients are not necessarily very sparse.

{We also run Shampoo in the same settings and we found that the memory usage is lower, but comes at a significantly higher running time (around $3\times$). We provide details on Shampoo comparison in Appendix~\ref{appendix:imagenet-rn18-shampoo}.}

To examine memory, we ran the experiments on a single A100 GPU and recorded the max memory usage throughout the entire process via the \textsc{nvidia-smi} tool. We find that SGD uses $12.8$ GB and S-MFAC $14.3$ GB, while D-MFAC in the same setting needs $59$ GB. 

\begin{figure}
\caption{\label{figure:imagenet-resnet18} Validation (Top-1) accuracy for ResNet18 training from scratch on ImageNet-1K. M-FAC and 99\%-Sparse M-FAC reach approximately 1\% higher validation accuracy relative to SGD using the well-tuned FFCV recipe~\cite{FFCV}. }
   \begin{minipage}{0.99\linewidth}
        \includegraphics[width=0.99\linewidth]{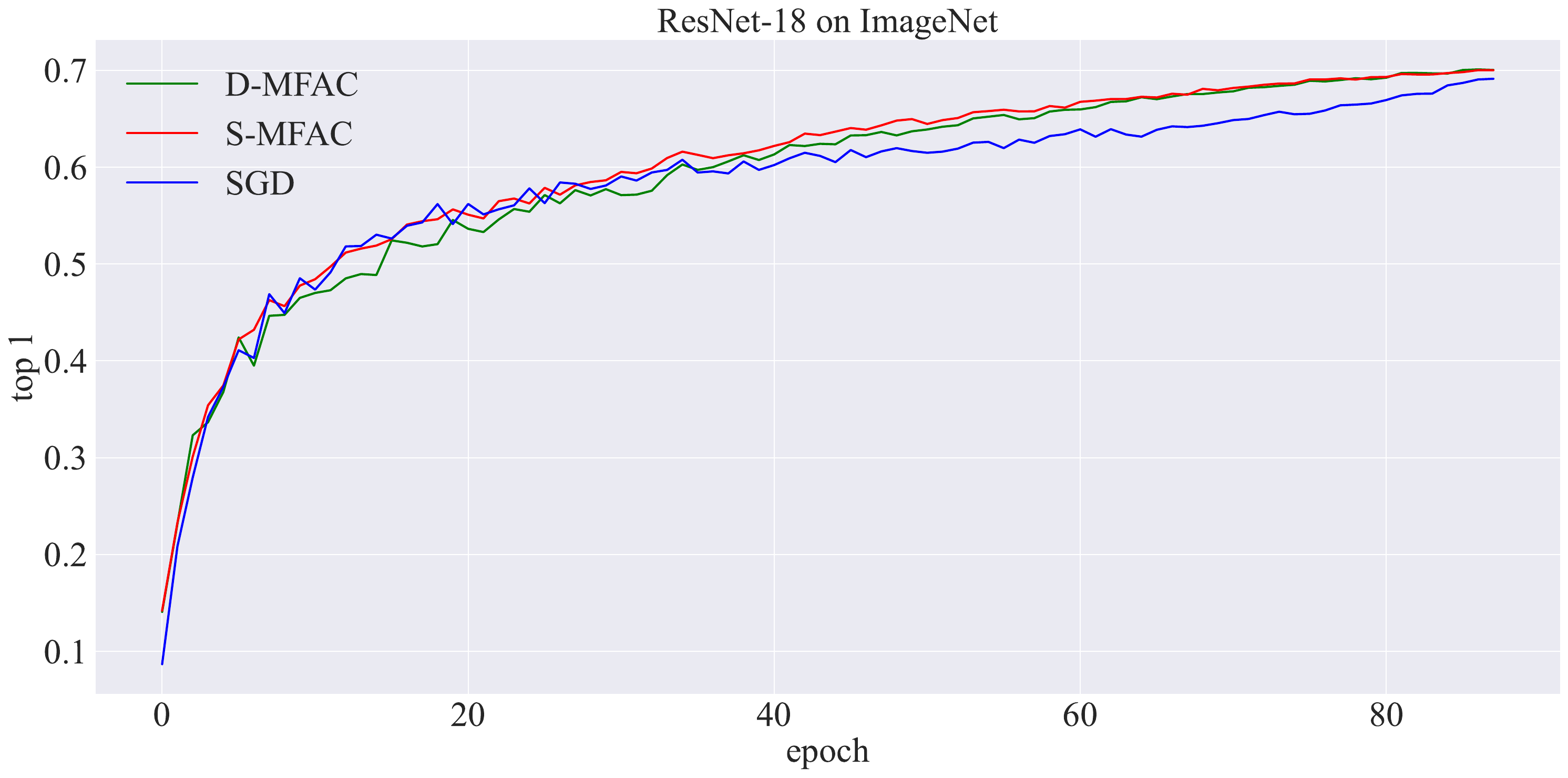}
    \end{minipage}
\end{figure}

\begin{figure}[!ht]
    \caption{\label{table:imagenet-resnet18} Training loss and validation (Top-1) accuracy for ResNet18 training from scratch on ImageNet-1K under a detailed sparsity sweep, between $1\%$ and $0.04\%$ density. Dense M-FAC outperforms SGD in this setting by about $1\%$, which is recovered by S-MFAC at 1\% density. Even with $0.17\%$ gradient density, M-FAC slightly outperforms SGD in terms of training loss and test accuracy. }
    \begin{minipage}{0.99\linewidth}
        \begin{center}
        \small
        \begin{tabular}{c c c c} 
            \toprule
            \textbf{Density $k$} & \textbf{Train Loss} & \textbf{Test Accuracy}\\
            \midrule
            D-MFAC (100\%) & 2.164 & 70.03\%\\
            \midrule
            \textbf{117k (1.00\%)} & \textbf{2.185} & \textbf{70.01\%} \\
            100k (0.85\%) & 2.183 & 70.18\%\\
            75k (0.64\%)& 2.187 & 69.99\%\\
            50k (0.42\%)& 2.190 & 69.68\%\\
            25k (0.21\%)& 2.206 & 69.43\%\\
            \textbf{20k (0.17\%)} & \textbf{2.217} & \textbf{69.26\%}\\
            15k (0.12\%)& 2.230 & 69.05\%\\
            10k (0.08\%)& 2.246 & 68.84\%\\
            5k (0.04\%)& 2.300 & 68.27\% \\
            \midrule
            SGD & 2.237 & 69.12\%\\
            \bottomrule
        \end{tabular}
        \end{center}
    \end{minipage}
\end{figure}

\paragraph{BERT/GLUE-MNLI.} Finally, we test our S-MFAC implementation for BERT-\textsc{Tiny}/\textsc{Mini}/\textsc{Base} models on the MNLI task from the GLUE benchmark~\cite{2019-wang-glue}, comparing against the Adam optimizer with optimized default settings from Hugging Face~\cite{2019-devlin}, which reproduce or improve upon the original paper~\cite{2019-devlin}. Dense M-FAC with a window size of $m=1024$ can only be executed on the \textsc{tiny} and \textsc{mini} model variants, requiring $18$ GB and $45$  GB memory, respectively, and runs out of memory (OOM) for the \textsc{base} model (the required memory usage for the ring buffer $G$ alone is 417GB).  

The results in Table~\ref{table:bert} show that our S-MFAC implementation not only recovers the performance of D-MFAC on BERT-\textsc{tiny}/\textsc{mini}, but also allows us to run on BERT-\textsc{base}, with superior validation accuracy relative to Adam using standard hyper-parameters (e.g., learning rate $\eta=2 \cdot 10^{-5}$). For the \textsc{tiny} model S-MFAC achieves almost $2\%$ higher accuracy compared to D-MFAC and close to $5\%$ higher validation accuracy compared to Adam. 
The former improvement can be justified by the regularization effect of sparsity relative to D-MFAC in a fine-tuning task, whereas the latter is due to leveraging second-order information.  

At the same time we observe that, accuracy differences become smaller as model size increases. This behavior is also reported in the original D-MFAC~\cite{2021-MFAC} optimizer for the same GLUE/MNLI task. While this is another indication that the sparse implementation  preserves the behavior of the dense counterpart, it suggests that the benefits of second-order information diminish in the case of over-parametrized fine-tuning tasks.

\begin{table}[!h]
\begin{center}
\caption{\label{table:bert} Evaluation accuracy for BERT-\textbf{T}iny/\textbf{M}ini/\textbf{B}ase on GLUE-MNLI. OOM stands for Out of Memory error.}
\begin{tabular}{ccccc}
    \toprule
                    & \textbf{T} & \textbf{M} & \textbf{B} \\
    \midrule
    \textbf{Adam}   & 66.86 & 74.77 & 84.6 \\
    \textbf{S-MFAC} & \textbf{71.06} & \textbf{76.40} & \textbf{84.86} \\
    \textbf{D-MFAC} & 69.26 & \textbf{76.39} & OOM \\
    \bottomrule
\end{tabular}
\end{center}
\end{table}

\paragraph{Running times and memory usage.} Finally, we examine running times and memory usage for the algorithm.  The results across our experiments are shown in Table~\ref{table:time-and-memory}. 
We note that, in the current implementation, both D-MFAC and our optimizer have a constant additional cost relative to the highly-optimized Adam or SGD implementations in Pytorch, due to the additional data structures employed. This cost can be reduced via optimizations.

Specifically, our current implementation was primarily optimized for residual networks. This is reflected in the runtime and memory numbers: on ResNet-18 ($11.7$M params), D-MFAC requires $4.6\times$ more memory compared to SGD with momentum and $25\%$ larger running time. Our S-MFAC implementation diminishes these costs \AK{bringing memory usage to $25\%$ of D-MFAC and making it comparable with} SGD both in the running time and memory usage.

For BERT-\textsc{tiny} ($4$M params), D-MFAC uses $14\times$ more memory and BERT-\textsc{mini} ($11$M params) uses $15\times$ more memory compared to S-MFAC. Further, we find that EFCP is extremely effective for large models like BERT-\textsc{Base} ($110$M params). In this case, running D-MFAC would raise an out of memory (OOM) error (just the dense ring buffer used to store the gradient window  requires 417 GB of memory under standard parameters). 

Although our primary concern in experiments is memory cost, we observe that, in terms of running time, our algorithm has essentially no overhead on ResNet-18. On BERT models, there is a notable runtime overhead, which is due in part to the fact that our CUDA kernels are not specifically adapted to BERT models. We expect this overhead to be reduced via tuning and further optimizations, which we plan to investigate in future work. However, we note that these single-GPU runtimes are very manageable already.

To decrease memory usage and running time, one can try reducing the number of gradients $m$. However, it is clear from prior work~\cite{agarwal2019efficient, 2021-MFAC} that large values of $m$ are needed to provide better accuracy results, and specifically to improve over standard baseline. This is theoretically motivated in part by the fact that the rank of matrix $G$ is at most $m$, which can be extremely low compared to the rank of Hessian matrix for large models and the approximation would be very poor for low values of $m$. Thus, large models require large number of gradients $m$, which renders our technique effective. To our knowledge, our approach allows full-matrix adaptive optimizers to be run on a consumer GPU for the first time. Appendix~\ref{appendix:memory-savings} contains a detailed theoretical analysis for memory savings.

\begin{table}[!hb]
\begin{center}
\caption{\label{table:time-and-memory} Running times and memory usages for ImageNet/ResNet-18 and GLUE/BERT reported on an NVIDIA A100 GPU with 82GB RAM. OOM stands for Out Of Memory.}
\small
\begin{tabular}{c c c c c c}
    \toprule
    \textbf{Model} & \textbf{Optimizer} & \textbf{Memory} & \textbf{Training Time}\\
    \midrule
               & SGD    &  12.8 GB & 4h 10m \\
     ResNet-18 & S-MFAC &  14.3 GB & 4h 21m \\
               & D-MFAC &  58.5 GB & 5h 33m \\
    \midrule
               & Adam   &  0.8 GB & 11m \\
     BERT-Tiny & S-MFAC &  1.3 GB & 1h 5m \\
               & D-MFAC & 17.9 GB & 1h 20m \\
    \midrule
               & Adam   &  1.3 GB & 21m \\
     BERT-Mini & S-MFAC &  2.9 GB & 1h 5m \\
               & D-MFAC & 44.9 GB & 1h 50m \\
    \midrule
                & Adam   &  7.1 GB & 1h 45m \\
     BERT-Base  & S-MFAC & 14.8 GB & 3h 20m \\
                & D-MFAC &   OOM   & -\\
    \bottomrule
\end{tabular}
\end{center}
\end{table}

\paragraph{Error Feedback Evolution During Training.}{In line 5 of Algorithm~\ref{algorithm:generic-efcp} the error feedback mechanism ensures that the top largest values by magnitude (the outliers) are always transferred at each step from the error buffer $\xi$ into the optimizer state and we are interested in whether the EF stays unbounded relative to the gradient values applied. To showcase this practically, we measured the size of the error feedback relative to the total gradient norm applied up to a certain step, formally $||\xi_t||_2 / (||g_1||_2 + … + ||g_t||_2)$. Intuitively, this should tell us whether the gradient information is stuck in the error feedback (equivalent to an increase or large constant value of this metric) or whether the error feedback acts like a temporary buffer before the information is transferred to the optimizer’s state (in which case it should become smaller over time). We plot this metric during the entire training process for two experiments, BERT-Base/GLUE-MNLI and ResNet-20/CIFAR-10, where we use $k=1\%$ gradient density for S-MFAC. We found that in both experiments this metric decreases exponentially, which indicates the accumulated gradient information is indeed transferred from the error feedback $\xi$ to the optimizer state. We show the experiment result in Appendix~\ref{appendix:ef-evolution}.}

\paragraph{Generalization.} {Since the preconditioner is highly sparse, we are interested in how our technique affects generalization. To address this, we inspected the eigenvalues of the loss landscape by conducting an experiment on ResNet-20 / CIFAR-10 to compute the maximum eigenvalue of the Hessian matrix of the loss function. We used the Power Iteration method for SGD, D-MFAC and S-MFAC with gradient sparsities 90\%, 95\%, 99\%, 99.5\% and 99.9\%. We use the entire train and test datasets and report the maximum eigenvalue after the initialization and then at the end of each epoch. We found that D-MFAC and S-MFAC yield much flatter solutions that SGD and S-MFAC yields only slightly larger maximum eigenvalues compared to D-MFAC. This is an indication that compressing the buffer yields the same optimizer properties. We provide the results in Appendix~\ref{appendix:generalization-eigenvalues}.}

\paragraph{Tuning Recipe.} {Our method has similar behavior as the original D-MFAC. The most important hyper-parameters are learning rate $\eta$, dampening $\lambda$, window size $m$ and batch size. Briefly, the batch size controls the noise in the gradient and D/S-MFAC have stable convergence for both large and small batch sizes; regarding dampening, we found some general rules of thumb for how to tune the dampening in relationship with the learning rate, based on the idea that changing dampening makes S/D-MFAC to behave like as SGD (for large values) or like Newton's method (for small values); regarding buffer size $m$, as explained in the original D-MFAc paper, it is recommended to use $m=1024$ for best results; regarding gradient density, we observed that there is a natural density threshold above which using higher density yields same results. In Appendix~\ref{appendix:tuning-recipe} we provide more details on how to tune these parameters to obtain the best out of our method.}

\section{Conclusions and future work}
We have provided a versatile new tool for addressing the memory costs of adaptive full-matrix preconditioning, 
in the form of a series of methods which compress the gradient history, by leveraging an instance of the error feedback mechanism.
We complement this mechanism via efficient sparse/low-rank algorithms for maintaining existing preconditioners, 
and have shown experimentally that our approach can essentially provide lossless compression when applied via sparsity, 
with remarkably large compression rates.
We provide CUDA implementations for a sparse data structure that 1) allows easy replacement of any row (adding/replacing new compressed representations of error feedback accumulators and 2) allows computing the SP and LCG operations efficiently which, to the best of our knowledge, is the first such data structure known in the literature.

In future work, we plan to further investigate theoretical justification for our method. Obtaining general bounds appears extremely challenging, as even in the case of standard SGD 
understanding error feedback required significant technical effort, e.g.~\cite{2018-alistarh, 2019-karimireddy}. 
Another direction of future work is exploring larger-scale validation of our method for additional architectures and tasks, and for additional preconditioning mechanisms. 

\section*{Acknowledgements}
The authors thank Adrian Vladu, Razvan Pascanu, Alexandra Peste, Mher Safaryan for their valuable feedback, the IT department from Institute of Science and Technology Austria for the hardware support and Weights and Biases for the infrastructure to track all our experiments.

\section*{Impact Statement}
This paper presents work whose goal is to improve certain aspects of existing algorithms in the field of Machine Learning. There are many potential societal consequences of our work, none which we feel must be specifically highlighted here because our work can be used by anyone without our control.

\bibliography{references}
\bibliographystyle{icml2024}

\newpage
\appendix
\onecolumn

\section{Error Feedback Evolution During Training}\label{appendix:ef-evolution}
{In this section we show the experimental results for the metric $||\xi_t||_2 / (||g_1||_2 + … + ||g_t||_2)$}

\begin{figure}[!h]
    \caption{\label{figure:ef-evolution-bert} Norms of the error feedback $||\xi_t||_2$, norm of the gradient $||g_t||_2$ and the EF metric value for BERT-Base.}
    \begin{minipage}{0.49\linewidth}
        \frame{\includegraphics[width=0.99\linewidth]{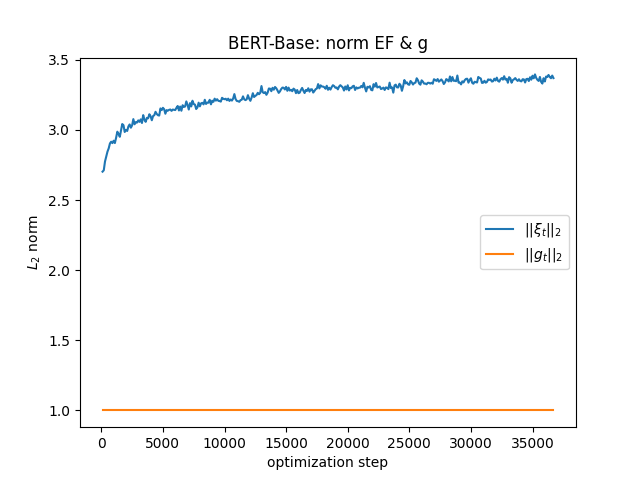}}
    \end{minipage}
    \begin{minipage}{0.49\linewidth}
        \frame{\includegraphics[width=0.99\linewidth]{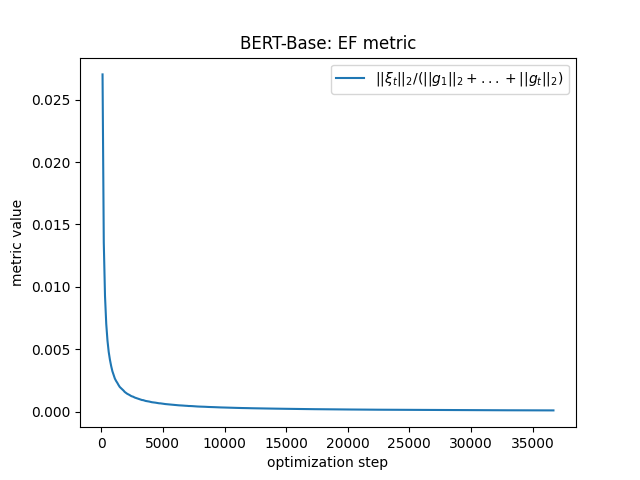}}
    \end{minipage}
\end{figure}

{In Figure~\ref{figure:ef-evolution-bert} for BERT, the stochastic gradient is clipped to norm 1 and the cumulative sum of the norm of the gradients up to step $t$ will always be $t$. In this case, the error feedback norm is much smaller than $t=||g_1|| + … + ||g_t||$, leading to an exponential decrease in the metric.}

\begin{figure}[!h]
    \caption{\label{figure:ef-evolution-resnet20} Norms of the error feedback $||\xi_t||_2$, norm of the gradient $||g_t||_2$ and the EF metric value for ResNet-20.}
    \begin{minipage}{0.49\linewidth}
        \frame{\includegraphics[width=0.99\linewidth]{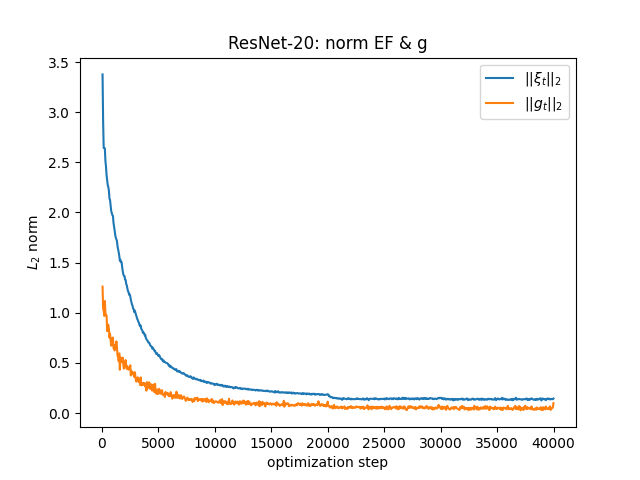}}
    \end{minipage}
    \begin{minipage}{0.49\linewidth}
        \frame{\includegraphics[width=0.99\linewidth]{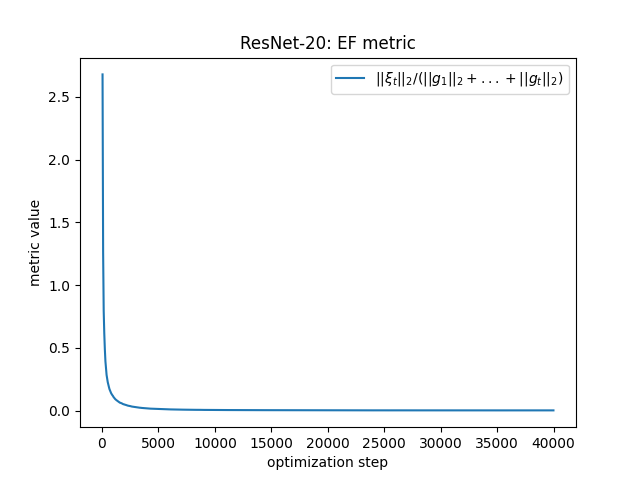}}
    \end{minipage}
\end{figure}

{In Figure~\ref{figure:ef-evolution-resnet20} for ResNet-20, the stochastic gradient is not clipped and its norm decreases naturally during the training. In this case, the error feedback norm decreases, also leading to an exponential decrease in the metric.}

{These two examples clearly indicate that the error feedback does not accumulate significant errors over time, but it acts as a temporary buffer of past gradient information before the top-k operator removes the outliers and are sent to the preconditioner.}

\section{Generalization.}\label{appendix:generalization-eigenvalues}
{In Table~\ref{table:generalization-eigenvalues} we show the results for our experiment on maximum eigenvalue of the hessian of the loss function. We observe that all M-FAC versions (dense and sparse) lead to a flat minima where the maximum eigenvalue is close to zero for both training and test datasets, while SGD leads to a sharp minimum with a maximum eigenvalue much larger than 0 for both training and test datasets. This behavior is similar to the one illustrated in Figure 1 of Keskar et al., available} \href{https://arxiv.org/pdf/1609.04836.pdf}{here}.

\begin{table}[!ht]
\begin{center}
\caption{\label{table:generalization-eigenvalues} Maximum eigenvalues computed on train / test datasets.}
\begin{tabular}{c c c c c c c c}
    \toprule
    \textbf{epoch} & \textbf{SGD} & \textbf{D-MFAC} & $\mathbf{k=0.1\%}$ & $\mathbf{k=0.5\%}$ & $\mathbf{k=1\%}$ & $\mathbf{k=5\%}$ & $\mathbf{k=10\%}$ \\
    \midrule
    init & \multicolumn{7}{c}{1231.34  /  942.1}  \\
    10 & 29.04 / 30.36 & 1.0 / 1.36 & 11.55 / 14.81 & 2.68 / 3.8 & 1.84 / 2.25 & 1.44 / 1.72 & 2.16 / 2.56 \\\
    20 & 17.37 / 29.46 & 0.22 / 0.23 & 1.55 / 1.83 & 0.38 / 0.43 & 0.27 / 0.34 & 0.26 / 0.33 & 0.35 / 0.42 \\
    40 & 14.61 / 20.84 & 0.09 / 0.11 & 0.25 / 0.39 & 0.11 / 0.14 & 0.09 / 0.1 & 0.08 / 0.11 & 0.11 / 0.14 \\
    60 & 10.38 / 21.34 & 0.05 / 0.09 & 0.13 / 0.23 & 0.05 / 0.11 & 0.05 / 0.1 & 0.05 / 0.08 & 0.06 / 0.11 \\
    80 & 10.19 / 24.51 & 0.05 / 0.1 & 0.11 / 0.25 & 0.07 / 0.12 & 0.06 / 0.1 & 0.05 / 0.1 & 0.06 / 0.11 \\
    100 & 9.12 / 24.51 & 0.04 / 0.1 & 0.1 / 0.23 & 0.07 / 0.1 & 0.06 / 0.11 & 0.05 / 0.11 & 0.05 / 0.12 \\
    \bottomrule
\end{tabular}
\end{center}
\end{table}

\section{ImageNet / ResNet-18 / Shampoo}\label{appendix:imagenet-rn18-shampoo}
{The ASDL experiments presented in Table~\ref{table:asdl-results} show that Shampoo is a strong competitor to our work with a lower memory usage. To address this, we integrated the PyTorch implementation of Shampoo optimizer from Google Research~\cite{gupta2018shampoo} into our ImageNet training pipeline. This reveals a few trade-offs when running Shampoo at scale. Specifically, its two key parameters are:}

\begin{itemize}
    \item {\textbf{preconditioning\_compute\_steps}: an integer that controls when the preconditioner is updated. For S-MFAC, this is always 1, since we re-compute at each step; however, Shampoo becomes very slow in this setting, so we had to increase this to 10 to run at reasonable speed.}
    \item {\textbf{block\_size}: an integer that controls the matrix approximation. We used the default 128, which means that we are performing a block diagonal approximation for the preconditioner.}
\end{itemize}

{The efficiency-accuracy trade-off for Shampoo is the following: a large block size makes Shampoo be closer to the more accurate full-matrix version, and should be paired with a large update interval. The configuration for Shampoo that makes it equivalent to D/S-MFAC would be to update the preconditioner at each step with maximum block size.}

{However, we were unable to run this “full” version of Shampoo at scale, due to speed/memory issues. For instance, the maximum block size we could run in an RTX 3090 GPU for ResNet18 was 8192 (3x slower than the default 128, and about 10x slower than S-MFAC in the same setting), and we had to reduce \textbf{preconditioning\_compute\_steps} to 10 to get reasonable training times on ImageNet.}

{We used Shampoo for training in exactly the same setting as in our paper, following the learning rates described in section 6.1 of the original paper [1]. To get reasonable running times, we re-computed the preconditioner once every 10 steps, and used the default suggested block size 128. We present our Shampoo results in Table~\ref{table:imagenet-rn18-shampoo}.}

\begin{table}[!ht]
\begin{center}
\caption{\label{table:imagenet-rn18-shampoo} Results for Shampoo / ImageNet / ResNet-18.}
\begin{tabular}{cccc}
    \toprule
    \textbf{lr} & \textbf{top1} & \textbf{time} & \textbf{memory usage} \\
    \midrule
    1   & 68.82\% & 11h 38m & 12.66GB \\
    2.5	& 69.62\% & 11h 19m & 12.66GB \\
    5   & 70.24\% & 11h 21m & 12.66GB \\
    \bottomrule
\end{tabular}
\end{center}
\end{table}

{In comparison, we report the results for our optimizer in the following table:}

\begin{table}[!ht]
\begin{center}
\caption{Results for S-MFAC / ImageNet / ResNet-18.}
\begin{tabular}{ccc}
    \toprule
    \textbf{top1} & \textbf{time} & \textbf{memory usage} \\
    \midrule
    S-MFAC (k=1\%) & 14.3 GB & 4h 21m \\
    \bottomrule
\end{tabular}
\end{center}
\end{table}

\begin{figure}[!h]
    \caption{\label{figure:imagenet-resnet18-shampoo} Comparison between S-MFAC and Shampoo on ResNet-18 / ImageNet}
    \begin{minipage}{0.49\linewidth}
        \frame{\includegraphics[width=0.99\linewidth]{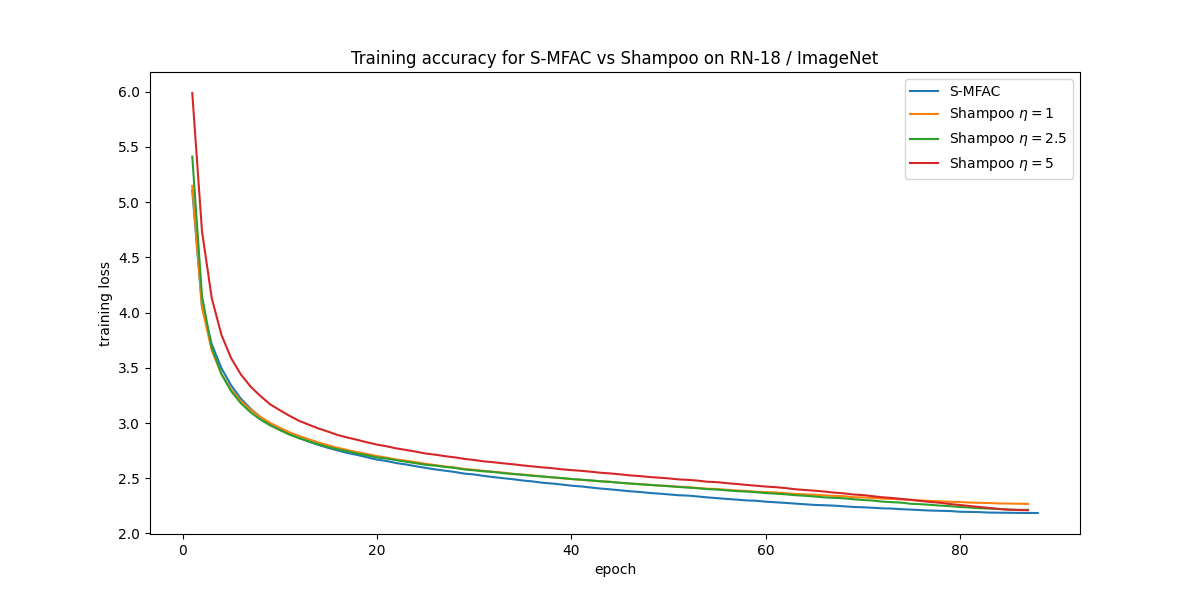}}
    \end{minipage}
    \begin{minipage}{0.49\linewidth}
        \frame{\includegraphics[width=0.99\linewidth]{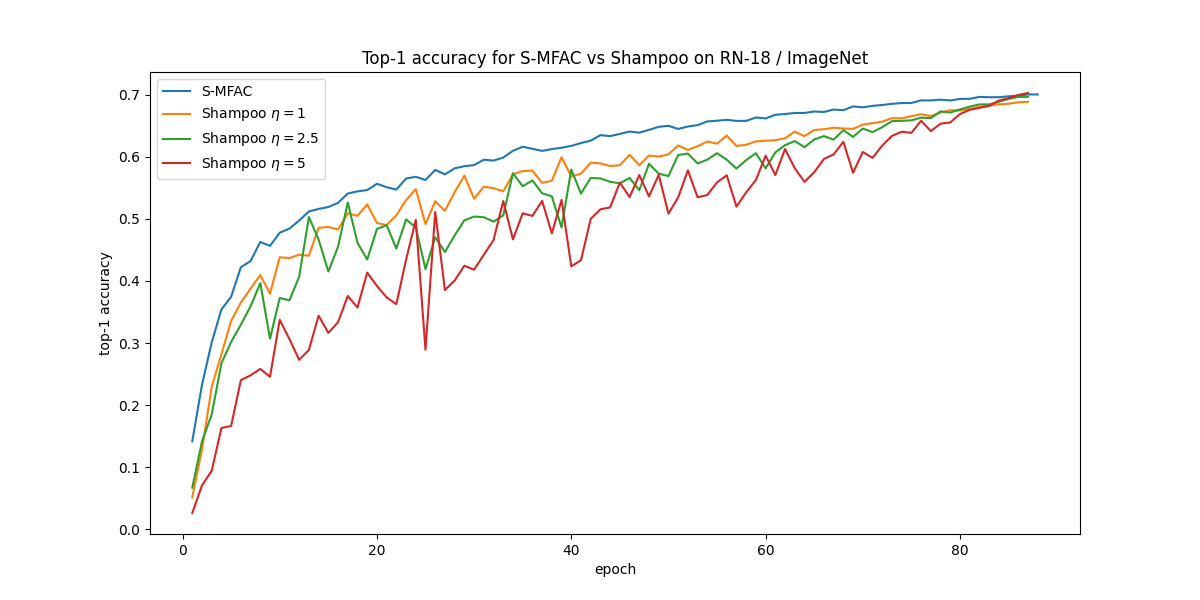}}
    \end{minipage}
\end{figure}

{We ran Shampoo in exactly the same settings as SGD, D-MFAC and our S-MFAC: linearly decaying learning rate, no image scaling, weight decay $1e-5$ for 88 epochs and batch size 1024 (1251 steps per epoch). We mention that some Shampoo runs crashed towards the end of training due to numerical instability, but we were still able to extract good-quality checkpoints. In Figure~\ref{figure:imagenet-resnet18-shampoo} we show the plots for top-1 accuracy and training loss for Shampoo.}

{We would like to mention that we also tried to train BERT-Base on GLUE/MNLI, but unfortunately the run was too slow. The total number of training steps is 36k and it took 2h 48m for 1200 steps. Since the model is large (110M params), we set the block size to 16384. A small block size (e.g. the default 128) would have resulted in a lot of smaller blocks that resulted in similar running times. Our S-MFAC approch takes 3h 20m for the entire run.}

\section{Tuning Recipe}\label{appendix:tuning-recipe}
{In this section we explain the influence of the main hyper-parameters.}

\paragraph{Batch size.} {The first parameter of interest is the batch size, which controls the noise in the gradient. We performed an ablation study on both S/D-MFAC for ResNet-20/CIFAR-10 (we use this small datasets/model for time reasons) without weight decay and with parameters that are reported by the original D-MFAC.}

{We tried batch sizes 32, 64, 128, 256 and 512. Both S/D-MFAC have stable convergence at both small and large batch sizes, although using a small batch size will result in a large total number of optimization steps and this might increase the running time (as it happens for any optimizer).}

{Using a large batch size for D-MFAC can lead to an Out of Memory Error (OOM) because both model and optimizer state have high memory footprint. In contrast, S-MFAC will keep a low memory footprint and one can increase the batch size accordingly such that the GPU memory usage is maximized.}

\paragraph{Dampening $\lambda$.} {Since M-FAC gives a Hessian approximation based on empirical Fisher, it is implicitly a trust region method. The dampening parameter $\lambda$ is inversely proportional to the unknown trust region radius $R$ and it controls the behavior of the optimizer. When $\lambda$ is small, M-FAC is similar to Newton's method, while large $\lambda$ reduces to SGD with a scaled learning rate $\eta_t / \lambda$.}
{We found two rules of thumb that proved to be useful in tuning the learning rate $\eta$ and dampening $\lambda$:}
\begin{enumerate}
    \item {if the model diverges, then the dampening is too small and it should be increased. This means S/D-MFAC is too close to Newton's method and by increasing it we make it more similar to SGD. The dampening influences the norm of the linear combination parameters $\mathbb{c}$ in Equation~\ref{formula:mfac-linear-combination}. If the norm becomes too large, this is equivalent to a behavior similar to Newton's method (too small dampening).}
    \item {if the model underfits, then the dampening is too large and it should be decreased. This means S/D-MFAC is too close to SGD and by decreasing it we make it more similar to Newton's method.}
\end{enumerate}

\paragraph{History size $m$.} {As explained in the original D-MFAC paper, it is recommended to use a large history size for the following reasons:}

{From a theoretical perspective, the rank of the Empirical Fisher Matrix (EFM) is at most $m$. This is already a limitation when approximating the real Hessian matrix (which is likely to have a rank $r >> m$) and using a value of $m$ as large as possible is recommended; from a practical perspective, the recommended value of $m$ is up to 1024. On the one hand, this is recommended since the maximum number of threads in a block (across which parallelization is performed) on recent NVIDIA GPUs Nvidia is 1024; on the other, we do not observe additional accuracy improvements for larger history sizes.}

\paragraph{Sparsity.} {We are using the same experiment on ResNet-20/CIFAR-10 to provide explanations about how we should set the sparsity. In our experiments so far we used the following density percentages $k$: 0.1\%, 0.5\%, 1\%, 5\% and 10\%. We have observed that there is always a natural density percentage threshold $k^*$ above which using higher density (equivalent to lower sparsity) yields the same results. For example, in our experiments so far we observed that $k=1\%$ obtains best results for $m=1024$ gradients, while for larger values of $k$ we observe a slight decline in test accuracy.}

{As a summary, we would suggest using $k=1\%, m=1024$ for the optimizer parameters, as “universal” parameters, which worked well across all our experiments. The user can choose the batch size to maximize memory/compute usage for the given GPU/TPU.}

\section{Hyper-parameters} \label{appendix:hyper-parameters}
We restate the notations that we introduced in the main body of our work. We use notation $\mu$ for momentum, $\eta$ for the initial learning rate, $\gamma$ for weight decay, $\lambda$ for M-FAC dampening, $k$ for gradient density, $\rho$ for rank, $E$ for epochs, $B$ for batch size.

\subsection{ASDL}
We perform hyper-parameter tuning for CIFAR-10 and ViT-Tiny, as described in the appendix section of the ASDL~\cite{ASDL} paper. Concretely, the general grids we use are $\{32, 128, 512, 2048\}$ for the batch size, $\{{3e-1,1e-1,3e-2,1e-2,3e-3,1e-3}\}$ for the learning rate, $E=20$ epochs, we clip the preconditioned gradient norm to $10$ and perform grid search for the dampening parameter $\lambda$ over the grid $\{1e-2, 5e-3, 1e-3, 5e-4, 1e-4, 5e-5, 1e-5, 5e-6, 1e-6\}$ for S-MFAC. In particular, ViT-Tiny uses $\gamma=1e-4$. 

\subsection{FFCV}
For the ImageNet dataset and ResNet-18 model, we use $E=88$ epochs, batch size $B=1024$, linear decay schedule for the learning rate $\eta$. In particular, we set the weight decay $\gamma=1e-5$ for convolution and fully connected parameters and $\gamma=0$ for batch-normalization parameters, thus fixing the issue in the main FFCV repository. We use linear learning rate decay for all three optimizers and disable image rescaling. We present the final hyper-parameters in Table~\ref{table:grid-ffcv}.

\subsection{BERT/GLUE}
We use the HuggingFace repository and plug in our D/S-MFAC optimizers. We compare against default Adam baselines as follows. For GLUE/MNLI we use $E=3, \eta=2e-5, \gamma=0$. D-MFAC uses the default parameters from the original paper, e.g. $\eta=1e-4, \lambda=1e-6, \gamma=0$ and we adopt the same values for S-MFAC for \textsc{BERT-Tiny/Mini}. For \textsc{BERT-Base} we use $\lambda=5e-5$ for S-MFAC.

\subsection{CIFAR-10/RN-20}
For D/S-GGT we use window size $m=100, k=1\%$ and the grid in Table~\ref{table:grid-synthetic-ggt}. We show the corresponding pairs $(\eta, \epsilon)$ for the best runs extracted from the search grids. In these experiments we use $E=164, B=128$ and decay the learning rate by a factor of $\gamma_\eta=0.1$ at 50\% and 75\% of training. The final hyper-parameters for each method are summarized in table~\ref{table:grid-cifar10-rn20}.

\begin{table}[!ht]
\begin{center}
\caption{\label{table:grid-synthetic-ggt} Search grid for learning rate $\eta$ and dampening $\epsilon$ for S/D-GGT.}
\small
\begin{tabular}{c c}
    \toprule
    $\eta, \epsilon$ & 1e-7, 1e-6, 1e-5, 1e-4, 1e-3, 1e-2, 1e-1, 1\\
    \midrule
    $\eta_D, \epsilon_D$ & loss: (1, 1e-6), accuracy: (1, 1e-5) \\
    \midrule
    $\eta_S, \epsilon_S$ & loss: (1, 1e-6), accuracy: (1, 1e-5) \\
    \bottomrule
\end{tabular}
\end{center}
\end{table}

\begin{table}[!ht]
\begin{center}
\caption{\label{table:grid-cifar10-rn20} Final hyper-parameters for CIFAR-10 / ResNet-20.}
\small
\begin{tabular}{c c}
    \toprule
    SGD & $\eta=0.1, \mu=0.9, \gamma=5e-4$ \\
    \midrule
    D-GGT & $m=100, \eta=0.1, \epsilon=1e-5, \gamma=0$ \\
    \midrule
    S-GGT & $m=100, \eta=0.1, \epsilon=1e-5, \gamma=0$ \\
    \midrule
    D-MFAC & $m=1024, \eta=1e-3, \lambda=1e-6, \mu=0, \gamma=1e-4$ \\
    \midrule
    S-MFAC & $k=1\%, m=1024, \eta=1e-3, \lambda=1e-4, \mu=0, \gamma=1e-4$ \\
    \midrule
    LR-MFAC & $\rho = 4, m=1024, \eta=1e-3, \lambda=1e-6, \mu=0, \gamma=0$ \\
    \bottomrule
\end{tabular}
\end{center}
\end{table}

\begin{table}[!ht]
\begin{center}
\caption{\label{table:grid-ffcv} Hyper-parameters for SGD and S/D-MFAC for FFCV/ImageNet.}
\small
\begin{tabular}{c c}
    \toprule
    SGD & $\eta=0.5, \mu=0.9$ \\
    \midrule
    D-MFAC & $\eta=1e-3, \lambda=1e-6, \mu=0$ \\
    \midrule
    S-MFAC & $\eta=1e-3, \lambda=1e-7, \mu=0$, $k=1\%$ \\
    \bottomrule
\end{tabular}
\end{center}
\end{table}

\section{Memory savings.} \label{appendix:memory-savings}
In this section we detail the theoretical memory savings for the sparse preconditioners by computing the total memory in bytes and the ratio of dense over sparse preconditioners to describe de memory savings.

\subsection{Sparse M-FAC vs Dense M-FAC}
For M-FAC we will consider that $m=1024$ and $k=d/100$.

Dense M-FAC stores a matrix $G \in \mathbb{R}^{m \times d}$ of 32 bit floats, having a total of $4md$ bytes, resulting in a theoretical memory footprint of $4096d$ bytes.

On the other hand, Sparse M-FAC stores multiple tensors:
\begin{itemize}
    \item $\mathcal{I} \in \mathbb{R}^{m \times k}$ of 32 bit integers for the indices
    \item $\mathcal{V} \in \mathbb{R}^{m \times k}$ of 32 bit floats or 16 bit bfloat16s for the values
    \item $\xi \in \mathbb{R}^{d}$ of 32 bit floats for the error feedback
    \item $r_m \in \mathbb{R}^{m}$ of 32 bit floats to store the SP result
    \item $r_d \in \mathbb{R}^{d}$ of 32 bit floats to store the LCG result
\end{itemize}

When $\mathcal{V}$ holds 16 bit bfloat16s, the memory footprint for $\mathcal{I}, \mathcal{V}, \xi, r_m, r_d$ altogether is $4mk + 2mk + 4d + 4m + 4d = 0.06md + 8d + 4m = 61.44d + 8d + 4m = 69.44d + 4m \approx 70d$.

When $\mathcal{V}$ holds 32 bit floats, the memory footprint for $\mathcal{I}, \mathcal{V}, \xi, r_m, r_d$ altogether is $4mk + 4mk + 4d + 4m + 4d = 0.08md + 8d + 4m = 81.92d + 8d + 4m = 89.92d + 4m \approx 90d$.

In the end, the memory savings for M-FAC is $\frac{4096d}{90d} \approx 45.5\times$ for float and $\frac{4096d}{70d} \approx 58.5\times$ for bfloat16.

\subsection{Sparse GGT vs Dense GGT}
For GGT we will consider that $m=100$ and $k=d/100$.

Dense GGT stores a matrix $G \in \mathbb{R}^{m \times d}$ of 32 bit floats, having a total of $4md$ bytes, resulting in a theoretical memory footprint of $400d$ bytes.

Sparse GGT stores the following tensors:
\begin{itemize}
    \item $\mathcal{I} \in \mathbb{R}^{m \times k}$ of 32 bit integers for the indices
    \item $\mathcal{V} \in \mathbb{R}^{m \times k}$ of 32 bit floats for the values
    \item $\xi \in \mathbb{R}^{d}$ of 32 bit floats for the error feedback
    \item $r_m \in \mathbb{R}^{m}$ of 32 bit floats to store the SP result
\end{itemize}

The theoretical memory footprint for $\mathcal{I}, \mathcal{V}, \xi, r_m$ altogether is $4mk + 4mk + 4d + 4m = 8mk + 4d + 4m = 0.08md + 4d + 4m = 8d + 4d + 4m = 12d + 4m \approx 12d$. In the end, the memory savings for GGT is $\frac{400d}{12d} \approx 33.3\times$.

\section{CUDA Kernels} \label{section:appendix-cuda-kernels}
In this section we describe our efficient CUDA kernels for large models in algorithmic form. We reiterate that M-FAC has two operations that have to be reimplemented when using a sparse ring buffer $G=(\mathcal{I}, \mathcal{V})$, the scalar products (SP) and linear combination of gradients (LCG). In the next paragraphs we provide a brief algorithmic description for SP and LCG.

\subsection{SP Kernel}
For this operation we have to compute the scalar product between each row in $G$ and a general gradient $g$. The sparse matrix $G$ is splitted into two matrices, the indices matrix $\mathcal{I}\in\mathbb{N}^{m \times k}$ and the values matrix $\mathcal{I}\in\mathbb{R}^{m \times k}$. We use 1024 thread blocks, each containing 1024 threads and one thread block will process one row. In the algorithm we will consider that $\mathcal{I}$ and $\mathcal{V}$ are 2-dimensional, but in practice they are 1-dimensional. We use $\omega \in \mathbb{R}^m$ to store result of the SP operation in Algorithm~\ref{algorithm:cuda-sp}

\begin{algorithm}[!ht]
    \caption{SP kernel (CUDA)}\label{algorithm:cuda-sp}
    \begin{algorithmic}[1]
        \STATE Input: $m, k, g, \mathcal{I}, \mathcal{V}, \omega$
        \STATE $mem$ is a shared memory buffer with $T$ components
        \STATE $\sigma \gets 0$ \COMMENT{$\triangleright$ stores the partial dot product in current thread}
        \STATE $Bid \gets $ block id \COMMENT{$\triangleright$ gives the row index in $\mathcal{I}$ and $\mathcal{V}$}
        \STATE $Tid \gets $ thread id
        \STATE $T \gets $ number of threads in the block
        \STATE $r \gets Bid$ \COMMENT{$\triangleright$ row index}
        \FOR{$j \in \{T_{id}<k,T\}$ \COMMENT{$\triangleright$ $j$ from $T_{id}$ to $k$ with step $T$ using coalesced memory access}}
            \STATE $i \gets \mathcal{I}_{r,j}$
            \STATE $v \gets \mathcal{V}_{r,j}$
            \STATE $\sigma \gets \sigma + v \cdot g_i$ \COMMENT{$\triangleright$ accumulate partial dot product}
        \ENDFOR
        \STATE $mem_{Tid} \gets \sigma$
        \STATE parallel\_reduce(mem, T) \COMMENT{$\triangleright$ logarithmic sum: result is in $mem_0$}
        \IF{$Tid = 0$}
            \STATE $\omega_r \gets mem_0$
        \ENDIF
    \end{algorithmic}
\end{algorithm}

\subsection{LCG Kernel}
The LCG operation is more complex than the SP operation and requires providing more details. We will start from the compression operator from line 6 of Algorithm~\ref{algorithm:generic-efcp}, which actually applies Top-$k$ on the accumulator $a_t$, returning a compressed representation $c_t$ (not to be confused with the vector $c$ that stores the coefficients of the linear combination) - the $c_t$ is placed on a specific row of $G=(\mathcal{I}, \mathcal{V})$ at each iteration $t$.

The Top-$k$ operator is applied in blocks of size $B_d$ (the subscript $d$ means that this block is part of a $d$-dimensional vector), which is chosen automatically to be the maximum number of floating point values that can fit in the entire shared memory space for a thread block. For example, the A100 GPU from Nvidia has CUDA capability 8.0 and the maximum amount of shared memory per block is 163KB, meaning that we can store $B_d=41728$ float32 values in the shared memory. The value $k$ is inputted as a percentage at runtime and represents the density percentage of $c_t$. We convert this percentage to an actual density in each block $B_d$. For example, for $k=1\%$ we will have around $B_k=B_d/100$ (we will further skip some implementation details).

The key detail of our LCG kernel is we use the shared memory of size $B_d$ to accumulate partial linear combinations over slices of size $(m, B_d)$ of the sparse matrix $G$ and then write the result to the output $\omega \in \mathbb{R}^d$. This is possible because we apply Top-$k$ operator in blocks of size $B_d$ and we know the ends of the interval where we write the output to. Each thread block can process multiple slices of size $(m, B_d)$, depending on how large the model is. The pseudocode of the LCG kernel is provided in Algorithm~\ref{algorithm:cuda-lcg}.

\begin{algorithm}[!ht]
    \caption{LCG kernel (CUDA)}
    \label{algorithm:cuda-lcg}
    \begin{algorithmic}[1]
        \STATE Input: $B_d, B_k, d, m, k, c, \mathcal{I}, \mathcal{V}, \omega$
        \STATE $B \gets $ total number of thread blocks
        \STATE $Bid \gets $ block id \COMMENT{$\triangleright$ gives the row index in $\mathcal{I}$ and $\mathcal{V}$}
        \STATE $Tid \gets $ thread id
        \STATE $T \gets $ number of threads in the block
        \STATE $mem$ is a shared memory buffer with $T$ components
        \STATE $mem \gets 0$ \COMMENT{$\triangleright$ initialize shared memory (the LCG accumulator) with zero}
        \STATE synchronize threads \COMMENT{$\triangleright$ wait for all threads to finish initializing}
        \STATE $n_{B_d} \gets \lceil\frac{d}{B_d}\rceil$ \COMMENT{$\triangleright$ how many blocks of size $B_d$ we have}
        \STATE $workload \gets \lceil\frac{n_{B_d}}{B}\rceil$ \COMMENT{$\triangleright$ how many blocks of size $B_d$ each thread block processes}
        \FOR{$i \in \{0<workload\}$ \COMMENT{$\triangleright$ iterate through all slices $(m, B_k)$ (workload) for the curent thread block}}
            \STATE $start_d \gets B_d \cdot B_{id} \cdot workload + B_d \cdot i$ \COMMENT{$\triangleright$ where slice starts in $g$}
            \STATE $end_d \gets min(d, start_d + B_d)$ \COMMENT{$\triangleright$ where slice ends in $g$}
            \STATE $start_k \gets B_k \cdot B_{id} \cdot workload + B_k \cdot i$ \COMMENT{$\triangleright$ where slice starts in $G$}
            \STATE $end_k \gets min(B_k, start_k + B_k)$ \COMMENT{$\triangleright$ where slice ends in $G$}

            \FOR{$row \in \{0<workload\}$ \COMMENT{$\triangleright$ iterate all rows in the slice $(m, B_k)$ between $(start_k, end_k)$}}
                \FOR{$col \in \{start_k+T_{id},min(end_k, end_d), T\}$ \COMMENT{$\triangleright$ jump $T$ steps (coalesced memory access)}}
                    \STATE $i \gets \mathcal{I}_{r,col}$ \COMMENT{$\triangleright$ read index}
                    \STATE $v \gets \mathcal{V}_{r,col}$ \COMMENT{$\triangleright$ read value}
                    \STATE $mem_{i-start_d} += c_r \cdot v$ \COMMENT{$\triangleright$ multiply the coefficient of current row with value}
                \ENDFOR\COMMENT{$\triangleright$ finished processing one row}
                \STATE synchronize threads
            \ENDFOR \COMMENT{$\triangleright$ finished processing all rows}
            \STATE \COMMENT{$\triangleright$ finished processing a slice of size $(m, B_k)$, now write result}
            \FOR{$i \in \{T_{id}<min(B_d, d-start_d),T\}$ \COMMENT{$\triangleright$ write result (coalesced memory access)}}
                \STATE $\omega_{i + start_d} \gets mem_i$ \COMMENT{$\triangleright$ write result from $mem$ to output $\omega$}
                \STATE $mem_i \gets 0$ \COMMENT{$\triangleright$ zerorize shared memory to prepare for the next step}
            \ENDFOR
            \STATE synchronize threads
        \ENDFOR
    \end{algorithmic}
\end{algorithm}

\section{Sparse GGT.} \label{section:appendix-sparse-ggt}
In principle, the same sparsity format can be used for GGT as in the case of M-FAC. However, we would like to provide a slightly different approach just to show that an existing sparsity format is not completely efficient with respect to memory since they store redundant information. Our Sparse GGT uses the same sparsity format as S-MFAC and here we want to emphasize some drawbacks of COO format.

At a high level, we wish to leverage sparsity in the logical gradient history matrix $G \in \mathbb{R}^{m \times d}$. For this, we store it in the COOrdinate list sparse format, which requires storing matrix $\mathcal{V} \in \mathbb{R}^{m \times k}$ for values, matrix $\mathcal{R} \in \mathbb{Z}^{m \times k}$ for rows and $\mathcal{C} \in \mathbb{Z}^{m \times k}$ for columns. All these matrices can be stored in vector format, with dimension $m \cdot k$. This will be compatible with standard sparse operations, such as the Sparse Matrix-Matrix (SPMM) method in the PyTorch Sparse library~\cite{PytorchSparse}.


We will use the tuple $(\mathcal{R}, \mathcal{C}, \mathcal{V})$ to refer to the sparse buffer $G$. We distinguish between $\mathcal{V}$, which is the internal state of the sparse buffer (holding the $k$ most significant values from $a_t$) and $\mathcal{V}_t$, which is the output of Top-$k$ operator at a single step. \textit{With slight abuse of notation, we call the compressed accumulator $c_t = (\mathcal{V}_t, \mathcal{I}_t)$ a sparse gradient, even though it is an accumulation of previous gradients.}
The matrix $\mathcal{C}$ is created in such a way that each row contains the index of that row (e.g. [0,...,0; 1,...,1;...; m-1,...,m-1]) and is represented as a single vector where the rows are concatenated (here is where we have redundancy, e.g. duplicated data - the values in this tensor never change).

Let $i \in \{0, ..., m-1\}$ be an index indicating the location where the new gradient will be placed in the buffer $(\mathcal{R}, \mathcal{C}, \mathcal{V})$. At each step, we receive the accumulator $a_t$, the most significant $k$ values $\mathcal{V}_t$ and their corresponding indices $\mathcal{I}_t$. The expression $a_t[\mathcal{I}_t] \in \mathbb{R}^d$ will represent a ``densified'' version of a sparse vector, which contains zeros at indices $j \notin \mathcal{I}_t$ (sparse tensor which also contains the zeros - does not provide any memory reduction).

We integrate the values and indices in the internal buffers $\mathcal{V}$ and $\mathcal{I}$ in row $i$ using the indices $\alpha$ and $\beta$. Since the sparse $G \in \mathbb{R}^{m \times k}$ and we have to compute the scalar product $\delta=G^Tx$, we can easily transpose the sparse $G$ by swapping the rows and columns in the SPMM call. The scalar products matrix $G^TG$ is updated by overwriting the $i^{th}$ row and column by $\delta$. At this point we can perform eigenvalue decomposition of $G^TG$ and finally compute the update direction $u_t$ by preconditioning the sparse accumulator $a_t[\mathcal{I}_t]$. 
The full procedure is presented in Algorithm~\ref{algorithm:full-sparse-GGT}.

\paragraph{GGT results for CIFAR-10/ResNet-20.} Our experimental setup covers ResNet-20~\cite{ResNet} ($d=0.27$M) for CIFAR-10~\cite{CIFAR10}. We train models from scratch using SGD with momentum, dense and sparse GGT (99\% sparsity) and M-FAC and low-rank M-FAC (rank $4$). We decouple the weight decay $\gamma$ from the gradient, and apply it in the optimizer, which allows us to compare the loss values. The results in Table~\ref{table:cifar10-resnet20} show validation accuracy on the ResNet20/CIFAR-10 experiment. We observe that 
1) Dense(D) M-FAC reaches similar accuracy to SGD, whereas Dense(D) GGT achieves lower accuracy in this task; 
2) Sparse(S) M-FAC reaches the same accuracy as the dense variant (and SGD); 
3) Low-Rank(LR) M-FAC with rank $4$ reaches < 0.8\% lower accuracy relative to the best results.

\begin{table}[!h]
\begin{center}
\caption{\label{table:cifar10-resnet20} Results for ResNet-20 on CIFAR10. The first row reports the training loss and second row validation accuracy. The number in parentheses indicate stdev (for training loss, this is within $0.001$).}
\small
\begin{tabular}{c c c c c c c}
    \toprule
    & \textbf{SGD} & \textbf{D-MFAC} & \textbf{S-MFAC} & \textbf{D-GGT} & \textbf{S-GGT} & \textbf{LR-MFAC} \\
    \midrule
    Loss & 0.032  & 0.017 & 0.017  & 0.108  & 0.097  & 0.021 \\

    Acc. & 92.16 ( .10) & 92.12 (.13) & 92.25  (.21) & 87.77 (.22) & 87.73 ( .34)  & 91.43  (.30)\\

    \bottomrule
\end{tabular}
\end{center}
\end{table}

\begin{algorithm}[!ht]
\caption{Detailed Sparse GGT Implementation}\label{algorithm:full-sparse-GGT}
\begin{algorithmic}[1]
\label{algorithm:sparse-ggt}
    \STATE Variables: $\mathcal{V} \gets 0 \in \mathbb{R}^{mk}$, $\mathcal{R} \gets 0 \in \mathbb{Z}^{mk}$, $i \gets 0$, $G^TG \gets 0 \in \mathbb{R}^{m \times m}$
    \STATE Operators: $\mathcal{C} \gets vec([r \cdot 1_k]_{r=0}^{m-1}) \in \mathbb{Z}^{mk}$ 
\newline
\STATE \textbf{function} \textsc{SparseGGT-Step}($t, a_t \in \mathbb{R}^d, \mathcal{V}_t \in \mathbb{R}^k, \mathcal{I}_t \in \mathbb{Z}^k$)
\STATE $\alpha \gets i \cdot k, \; \beta \gets \alpha+k, \; \mathcal{V}_{\alpha:\beta} \gets \mathcal{V}_t, \; \mathcal{R}_{\alpha:\beta} \gets \mathcal{I}_t$ \COMMENT{$\triangleright$ integrate values and indices}
\STATE $\delta \gets \textsc{SPMM}(Sparse(r=\mathcal{C}, c=\mathcal{R}, v=\mathcal{V}), Dense (a_t[\mathcal{I}_t]))$ \COMMENT{$\triangleright$ compute $G^Ta_t[\mathcal{I}_t]$}
\STATE $row_i(G^TG) \gets \delta$ \COMMENT{$\triangleright$ update dot products in row and col $i$}
\STATE $col_i(G^TG) \gets \delta$ \COMMENT{$\triangleright$ update dot products in row and col $i$}
\STATE $G^TG \gets V \Sigma_m^2 V^T$ \COMMENT{$\triangleright$ eigenvalue decomposition}
\STATE $i \gets (i+1) \; \mod \; m$ \COMMENT{$\triangleright$ index update}
\STATE $U_m \gets \textsc{SPMM}(Sparse(r=\mathcal{R}, c=\mathcal{C}, v=\mathcal{V}), Dense (V \sqrt{\Sigma_m}^\dagger)$
\STATE $u_t \gets \frac{1}{\epsilon} (a_t[\mathcal{I}_t]) + U_m\left[(\Sigma_m+\epsilon I_m)^{-1} - \frac{1}{\epsilon}I_m\right]U_m^T (a_t[\mathcal{I}_t])$ \COMMENT{$\triangleright$ compute GGT direction}
\STATE \textbf{return} $u_t$
\end{algorithmic}
\end{algorithm}

\section{Compressing Preconditioners via Low-Rank Approximation and Error Feedback} 
\label{app:low-rank}

In this section, we discuss an alternative approach for compressed preconditioners, via low-rank compression of the gradients. 
Specifically, we will implement the $\textsc{Compress}$ step in Algorithm~\ref{algorithm:generic-efcp} via a low-rank approximation of the gradient, 
using an efficient variant of power iteration~\cite{vogelsPowerSGDPracticalLowRank2019}. 
Then, we consider the M-FAC algorithm, and reformulate its operations to leverage low-rank matrix operations for memory savings. 
We start by describing how the algorithm works a single layer, and then present the global algorithm, which computes preconditioning for the whole model.  

\paragraph{Low-Rank M-FAC for a Single Layer.} 
Assume that the gradient at a given layer of the model is represented by the $s$-dimensional tensor $g\in \mathbb{R}^{p_1 \times p_2 \times \ldots \times p_s}$. To compress the tensor, we use a variant of power iteration~\cite{vogelsPowerSGDPracticalLowRank2019} by firstly unfolding the tensor into a matrix $\Tilde{g} \in \mathbb{R}^{p_1 \times p_{2:s}}$, where $p_{2:s} = \prod_{i=2}^{s}p_i$. This matrix is then iteratively multiplied by a rank $\rho$ matrix $Q \in \mathbb{R}^{p_{2:s} \times \rho}$, 
which is reused from the previous iteration, to obtain the left decomposition matrix $P \in \mathbb{R}^{p_1 \times \rho}$. After orthogonalization of $P$, the updated right matrix $Q$ is obtained from the relation $\Tilde{g} = PQ^T$. Notice that $P$ and $Q$ have $(p_1 + p_{2:s})\rho$ elements which can be significantly smaller than $p_1 \cdot p_{2:s}$ for smaller ranks $\rho$. This compression procedure is outlined in Algorithm \ref{algorithm:powersgd}.

In order to ``introduce'' the current gradient $g$ into the M-FAC preconditioner data structure, 
we firstly need to compute an inner product between previously stored gradients and $g$ (see Equation~\ref{formula:mfac-linear-combination}). 
In matrix representation the product between two gradients corresponds to the Frobenius inner product. 
Specifically, for a given low-rank representation of gradient matrices $\Tilde{g_i} = P_iQ_i^T$ and  $\Tilde{g_j} = P_jQ_j^T$, 
the Frobenius product can be written as $\Tr(\Tilde{g_i}^T\Tilde{g_j}) = \Tr(Q_iP_i^TP_jQ_j^T) = \Tr((P_j^TP_i)^T(Q_j^TQ_i))$. 
Notice that matrix products $P_j^TP_i$ and $Q_j^TQ_i$ are only of size $\rho$ by $\rho$, so we compute them explicitly and then perform summation over their element-wise products to get the final result. 
As full gradients do not appear in this expression, we store just left and right gradients $P_i$ and $Q_i$ and update the M-FAC inner product matrix as needed.

The final preconditioned gradient update is computed as a weighted sum of previous gradients. The weights are computed via dynamic programming on top of the inner product values as in the original M-FAC. The final weighted sum of gradients is obtained by sequentially adding gradients reconstructed from their low-rank representation $g_{i} = P_jQ_j^T$.   

\begin{algorithm}[!ht]
\begin{algorithmic}[1]
\caption{Power iteration compression algorithm~\cite{vogelsPowerSGDPracticalLowRank2019}}\label{algorithm:powersgd}
\STATE \textbf{PowerCompress}(gradient tensor $g$, previous $Q$) \COMMENT{$\triangleright$ applied for each layer}
\STATE $\Tilde{g} \gets g.\textsc{Reshape}(p_1, p_{2:s})$
\STATE $P \gets \Tilde{g}Q$
\STATE $P \gets \textsc{Orthogonalize(P)}$
\STATE $Q \gets \Tilde{g}^TP$
\STATE \textbf{return} $(P, Q)$
\end{algorithmic}
\end{algorithm}

\begin{algorithm}[!ht]
\begin{algorithmic}[1]
\caption{Low Rank M-FAC preconditioning}\label{algorithm:lrmfac}
    \STATE INPUT: per-layer compressed (($P^1$, $Q^1$),  \ldots ,($P^L$, $Q^L$)) gradients
        \STATE STATE: per-layer ring buffers $G^\ell$ with compressed gradient history; full inner products $G^TG$ 
\STATE \textbf{Low Rank M-FAC}($P^1$, $Q^1$),  \ldots ,($P^L$, $Q^L$)
\STATE $\delta \gets $ zero vector of size $m$ \COMMENT{$\triangleright$ new full inner product}
\FOR{each layer $\ell \in \{1, 2, ..., L\}$ \COMMENT{$\triangleright$ aggregate inner products across the whole model}}
    \STATE Update $G^\ell$ with current gradient $(P^\ell, Q^\ell)$
    \FOR{each time step $\tau \in \{t - m + 1, ..., t\}$}
        \STATE $\delta_{t - \tau} \gets$ low-rank inner product $\Tr((g_{\tau}^{\ell})^T (P^\ell(Q^\ell)^T))$
    \ENDFOR        
\ENDFOR
\STATE Update row and column in $G^TG$ from $\delta$
\STATE \textbf{M-FAC:} compute summation weights $w \in \mathbb{R}^{m}$ from $G^TG$
\FOR{each layer $\ell \in \{1, 2, ..., L\}$}
    \STATE $u_t^\ell \gets  \epsilon^{-1}P^\ell(Q^\ell)^T + \sum_{\tau = t - m + 1}^{t} w_{\tau}g_{\tau}$ \COMMENT{$\triangleright$ aggregate gradients from the ring buffer}
\ENDFOR
\STATE \textbf{return} vector of per-layer updates $u_t$
\end{algorithmic}
\end{algorithm}

\paragraph{Global Low-Rank M-FAC.} 
To obtain an equivalent algorithm to the original M-FAC, independent per-layer steps are not sufficient as we would lose cross-layer correlation information. Instead, we use linearity of the inner product to compute the final result efficiently. 
In the original version of M-FAC, a current complete gradient from a $L$-layered model is represented as a concatenation of flattened per-layer gradients $\widehat{g} = \begin{pmatrix}\widehat{g}^1 & \widehat{g}^2 & \ldots & \widehat{g}^L\end{pmatrix}^T$, where $\hat{g}^i$ is the gradient from layer $i$. In this, representation an inner product between two gradients $\widehat{g}_i$ and $\widehat{g}_j$ can be expressed in a block form as $\left(\widehat{g}_i, \widehat{g}_j\right) = \sum_{\ell=1}^{L}\left(\widehat{g}_i^\ell, \widehat{g}_j^\ell\right)$. This dot product is then used to update the inner product matrix $G^TG$. 

In the context of our efficient scheme for single-layer low-rank update, we replace each inner product block by its low-rank counterpart to obtain the equivalent values. Recall that the computation of summation weights in the M-FAC needs access only to these inner products, so this part of the algorithm remains the same. The final gradient update is computed per-layer and we rely on linearity of the sum again to share the summation weights and dampening coefficient $\epsilon$ between layers. For more details refer to the pseudocode in Algorithm \ref{algorithm:lrmfac}.

\paragraph{Results for Low-Rank M-FAC.} We present and discuss the experiments for Low-Rank M-FAC on CIFAR-10/ResNet-20 in Table~\ref{table:cifar10-resnet20} and hyper-parameters in Appendix~\ref{appendix:hyper-parameters}. Next, we present an ablation study of low rank decomposition on both CIFAR-10/ResNet-20 and GLUE/MNLI for BERT-\textsc{tiny}.

To determine how the size of the decomposition rank $\rho$ affects the effectiveness of the low rank M-FAC algorithm, we perform a series of experiments on ResNet-20 and BERT models. For ResNet-20 we use CIFAR-10 dataset and the same parameters as in the main experiments. For BERT family we use \textsc{tiny}, \textsc{mini} and \textsc{base} variants and evaluate them on MNLI task from GLUE benchmark. For both setups we vary the rank from $1$ to $8$ and report the final test accuracy in the Table \ref{table:lowrank-ablation}. 

We observe that the decomposition rank significantly affects smaller BERT models but barely influences BERT-\textsc{base} and RN-20 model results. While the exact reason for such behaviour is not obvious, we hypothesise that for RN-20 the decomposition is effective even at rank $1$ due to gradient tensors already having a small first dimension, resulting in small compression error. We also note that Rank $8$ BERT-\textsc{base} run diverged, requiring more investigation into the stability of the training.

\begin{table}[!ht]
\begin{center}
\caption{Test accuracy on CIFAR-10 dataset for ResNet-20 (RN-20) and on MNLI for BERT-\textsc{tiny}, -\textsc{mini} and -\textsc{base}, trained with Low-Rank M-FAC with different decomposition ranks $\rho$.}
\label{table:lowrank-ablation}
  \begin{tabular}{ccccc}
  \toprule
rank $\rho$ & RN-20 & TINY & MINI & BASE \\
\midrule
1           &  91.40  &  57.34 & 66.88  &  84.42 \\
2           &  91.19  &  63.49 & 68.76  &  84.44 \\
4           &  91.43  &  66.44 & 70.93  &  84.45 \\
8           &  91.13  &  67.90 & 72.22  &   ---  \\
\bottomrule
\end{tabular}  
\end{center}

\end{table}

\section{Rank and Density Connection for Similar Memory Usage}\label{appendix:rank-topk-connection}
{To apply a low rank matrix decomposition we cast s-dimensional gradient tensor at each layer of the model into a matrix by keeping the first dimension intact and flattening out the others. The choice of rank to achieve equivalent memory savings is dependent on the model structure as each layer dimensions are different.}

{We would like to provide a theoretical approach to compute the corresponding gradient density $k=\Delta d$ ($d$ is the model size) such that S-MFAC has the same theoretical memory footprint as the LR-MFAC with rank $\rho$ fixed to a specific value, such as one in ${1, 2, 4, 8, 16}$. Let $n$ be the total number of layers, which are denoted by $L_1, ..., L_n$ and $p_1^{(i)},...,p_{n_i}^{(i)}$ be the dimensions of $L_i$, where $n_i$ is the number of dimensions for $L_i$. Moreover, the total number of elements in $L_i$ is $\prod_{j=1}^{n_i}p_j^{(i)}$. For example, if $L_1 \in \mathbb{R}^{100 \times 200 \times 300}$, we will have $n_i=3$ and $(p_1^{(1)}, p_2^{(1)}, p_3^{(1)}) = (100, 200, 300)$ and the total number of elements is $100 \cdot 200 \cdot 300 = 6M$.}

{Note that the model size $d=\sum_{i=1}^n \prod_{j=1}^{n_i}p_j^{(i)}$. As described in Appendix E of the paper, LR-MFAC with rank $\rho$ reduces the size of $L_i$ from $\prod_{j=1}^{n_i} p_j^{(i)}$ to $(p_1^{(i)} + \prod_{j=2}^{n_i}p_j^{(i)})\cdot\rho$. By extrapolating this to all layers, the total size of a full gradient $g_t \in \mathbb{R}^d$ will be compressed to the size $\rho\cdot\sum_{i=1}^n (p_1^{(i)} + \prod_{j=2}^{n_i}p_j^{(i)})$. Suppose we have already set $\rho$ for LR-MFAC and we would like to determine the corresponding fraction $\Delta$ of weights such that the gradient density $k=\Delta \cdot d$ for S-MFAC such that LR-MFAC and S-MFAC require the same amount of memory. \begin{equation*} \Delta \cdot d = \rho \left[\sum_{i=1}^n \left(p_1^{(i)} + \prod_{j=2}^{n_i}p_j^{(i)}\right)\right] \implies \Delta = \frac{\rho}{d}\sum_{i=1}^n \left(p_1^{(i)} + \prod_{j=2}^{n_i}p_j^{(i)}\right) \end{equation*} We apply this analysis to ResNet-20 (272k params) with 10 classes and present the values in Table \ref{table:rank-topk-connection}. We can see that if we set $\Delta=2.98\%$, S-MFAC will have the same memory usage as LR-MFAC with $\rho=1$. The experiments in the main body of the paper show that we can reach good results using $\Delta = 1\%$, which results in a lower memory footprint than LR-MFAC. Table 8 in Appendix shows an ablation on rank $\rho$ for ResNet-20 and BERT models.}

{In the Table~\ref{table:rank-topk-connection} below we show the optimal $k=\Delta \cdot d$ such that LR-MFAC with rank $\rho$ has the same memory footprint with S-MFAC.}

\begin{table}[!ht]
\begin{center}
    \caption{The rank $\rho$ for Low-Rank M-FAC and corresponding gradient density $\Delta$ for Top-$\Delta$ M-FAC such that both methods have similar memory usage.}
    \label{table:rank-topk-connection}
    \begin{tabular}{cccccc}
    \toprule
    $\rho$ & 1 & 2 & 4 & 8 & 16 \\
    $\Delta$ & 2.98\% & 5.97\% & 11.93\% & 23.86\% & 47.72\% \\
    \bottomrule
    \end{tabular}
\end{center}
\end{table}

\section{Theoretical Discussion}
\label{appendix:theory}

In this section, we discuss theoretical justifications for the proposed method. 

\paragraph{The Diagonal Case.} First, in the \emph{diagonal case}, described in Algorithm~\ref{algorithm:error-feedback-adagrad}, we can adapt a recent argument by Li, Karimi, and Li~\cite{li2022distributed} to show that the method should guarantee standard rates of $O( 1 / \sqrt{T} + \sigma^2 / \sqrt T + d / T )$ under standard assumptions on the objective, that is 1) smoothness; 2) unbiased and bounded stochastic gradients; and 3) bounded variance. 
Specifically, obtaining this result follows by specializing Corollary 1 in~\citet{li2022distributed}, specifically adapting their AMSGrad instance to Algorithm~\ref{algorithm:error-feedback-adagrad}, and adjusting the constants correspondingly.

\begin{algorithm}[!ht]
    \caption{Diagonal AdaGrad with Error Feedback}\label{algorithm:error-feedback-adagrad}
    \begin{algorithmic}
        \STATE $\xi_0 = 0_d \in \mathbb{R}^d$ \COMMENT{$\triangleright$ initialize error}
        \STATE $G_0 = 0_d\in \mathbb{R}^{d \times d}$ \COMMENT{$\triangleright$ initialize diagonal}
        \FOR{each step $t \in \{1, 2, ... T\}$}
            \STATE $g_t \gets \nabla_\theta L(\theta_t)$
            \STATE $a_t \gets \xi_{t-1} + g_t$ \COMMENT{$\triangleright$ add gradient to previous error}
            \STATE $c_t \gets \textsc{TopK}(a_t)$ \COMMENT{$\triangleright$ compress the accumulator}
            \STATE $\xi_t \gets a_t - c_t$ \COMMENT{$\triangleright$ update the error}
            \STATE $G_t \gets G_{t-1} + diag(c_t^2)$ \COMMENT{$\triangleright$ update diagonal}
            \STATE $\theta_{t+1} \gets \theta_t -\eta_t diag(G_t)^{-1/2} c_t$
        \ENDFOR
    \end{algorithmic}
\end{algorithm}

\paragraph{Discussion of the Full-Matrix Case.} 
In the highly-complex full-matrix case, further investigation is necessary to provide a full theoretical characterization of convergence. We expect that this will be very challenging since even the analysis of standard SGD with error feedback required significant technical advances~\cite{2018-alistarh, 2019-karimireddy}, and generally the convergence of full-matrix preconditioner variants is less well understood. 

Specifically, our discussion will consider the convergence of an ``intermediate'' version of the Sparse GGT algorithm, which we call Scaled Sparse GGT. 
The only difference from the Sparse GGT algorithm in Section~\ref{subsection:efcp-top-k} is the fact that we perform rescaling of the accumulated dense descent direction  $a_t$ before and after taking the Top-$k$ operation. 
(Intuitively, this is required so that the accumulator $a_t$ is considered in the ``correct basis'' at the current iteration, as per the standard AdaGrad analysis.)
Experimentally, we find that the practical convergence of the scaled and un-scaled variants is nearly identical in the diagonal case.   

 More precisely, denote by $H_t$ the Sparse M-FAC approximation for the preconditioner at step $t$. 
 Then, assuming we are interested in an  implementation of full-matrix Adagrad, we let the scaled Top-$k$ operator $T_{H_{t}}$ be 
defined as
\[
T_{H_{t}}\left(v\right)=H_{t}^{1/2}\text{TopK}\left({H_{t}}^{-1/2}v\right). 
\]

Assume that we use this scaled Top-$k$ operator instead of the standard one in the GGT implementation. 
We begin with the following two technical results, whose proofs are immediate. 
\begin{lemma}
\label{lem:topkbasis} Given a PSD matrix $H$, let the operator $T_{H}$
defined as
\[
T_{H}\left(v\right)=H^{1/2}\text{TopK}\left(H^{-1/2}v\right)
\]
Then it always holds that
\[
\left\Vert T_{H}\left(v\right)\right\Vert _{H^{-1}}^{2}\geq\frac{k}{n}\left\Vert v\right\Vert _{H^{-1}}^{2}
\]
\end{lemma}

The second technical lemma will allow  us to use an
appropriate potential function.
\begin{lemma}
Let $A$ be a PSD matrix and consider a rank-2 update $A'=A+uu^{\top}+vv^{\top}$.
Then it always holds that
\[
\ln\det\left(A'\right)\geq\ln\det\left(A\right)+\ln\left(1+\frac{1}{2}\left\Vert u\right\Vert _{A^{-1}}^{2}+\frac{1}{2}\left\Vert v\right\Vert _{A^{-1}}^{2}\right).
\]
\end{lemma}

Next, we can re-write our algorithm's iterations as follows: 
\begin{align*}
x_{t+1} & = x_{t}-H_{t}^{-1}{T_{H_{t}}\left( m_{t}+g_{t}\right)} \simeq x_{t}-H_{t}^{-1}\underbrace{T_{H_{t}}\left(H_{t}H_{t-1}^{-1}m_{t}+g_{t}\right)}_{\widetilde{g}_{t}}\\
m_{t+1} & = \left( m_{t}+g_{t} \right)-{T_{H_{t}}\left( m_{t}+g_{t}\right)} \simeq \left(H_{t}H_{t-1}^{-1}m_{t}+g_{t}\right)-\underbrace{T_{H_{t}}\left(H_{t}H_{t-1}^{-1}m_{t}+g_{t}\right)}_{\widetilde{g}_{t}}, 
\end{align*}
where in the approximation we only needed to assume that the Hessian estimate is stable from one iteration to another, i.e. that $H_{t}H_{t-1}^{-1}m_{t} \simeq m_t$.

In this context, our Hessian approximation is updated via the rule: 
\begin{align*}
H_{0} & =\delta I\\
H_{t} & = H_{t-1}+ 2 \eta {T_{H_{t}}\left( m_{t}+g_{t}\right)} {T_{H_{t}}\left( m_{t}+g_{t}\right)}^{\top}  \simeq  H_{t-1}+ \eta\widetilde{g}_{t}\widetilde{g}_{t}^{\top}+\eta T_{H_{t}}\left(g_{t}\right)T_{H_{t}}\left(g_{t}\right)^{\top}.
\end{align*}

Under these approximations, needed to consider the correct ``basis'' for the updates, we can examine the change in potential function involving the accumulation:

\begin{align*}
\frac{1}{2}\left\Vert \left(x_{t+1}-H_{t}^{-1}m_{t+1}\right)-x^{*}\right\Vert _{H_{t}}^{2} & =\frac{1}{2}\left\Vert \left(\left(x_{t}-H_{t}^{-1}\widetilde{g}_{t}\right)-H_{t}^{-1}\left(H_{t}H_{t-1}^{-1}m_{t}+g_{t}-\widetilde{g}_{t}\right)\right)-x^{*}\right\Vert _{H_{t}}^{2}\\
 & =\frac{1}{2}\left\Vert \left(x_{t}-H_{t-1}^{-1}m_{t}\right)-x^{*}-H_{t}^{-1}g_{t}\right\Vert _{H_{t}}^{2}\\
 & =\frac{1}{2}\left\Vert \left(x_{t}-H_{t-1}^{-1}m_{t}\right)-x^{*}\right\Vert _{H_{t}}^{2}+\frac{1}{2}\left\Vert g_{t}\right\Vert _{H_{t}^{-1}}^{2}-\left\langle g_{t},\left(x_{t}-H_{t-1}^{-1}m_{t}\right)-x^{*}\right\rangle 
\end{align*}
from where we conclude that
\begin{align*}
\left\langle g_{t},x_{t}-x^{*}\right\rangle  & =\frac{1}{2}\left\Vert \left(x_{t}-H_{t-1}^{-1}m_{t}\right)-x^{*}\right\Vert _{H_{t}}^{2}-\frac{1}{2}\left\Vert \left(x_{t+1}-H_{t}^{-1}m_{t+1}\right)-x^{*}\right\Vert _{H_{t}}^{2}+\frac{1}{2}\left\Vert g_{t}\right\Vert _{H_{t}^{-1}}^{2}+\left\langle g_{t},H_{t-1}^{-1}m_{t}\right\rangle 
\end{align*}
where we may use the upper bound 
\begin{align*}
\left\langle g_{t},H_{t-1}^{-1}m_{t}\right\rangle  & \leq\frac{1}{2}\left\Vert g_{t}\right\Vert _{H_{t-1}^{-1}}^{2}+\frac{1}{2}\left\Vert m_{t}\right\Vert _{H_{t-1}^{-1}}^{2}\leq\frac{d}{k}\left(\frac{1}{2}\left\Vert \widetilde{g}_{t}\right\Vert _{H_{t-1}^{-1}}^{2}+\frac{1}{2}\left\Vert T_{H_{t}}\left(g_{t}\right)\right\Vert _{H_{t-1}^{-1}}^{2}\right)
\end{align*}
for which the latter inequality follows from Lemma \ref{lem:topkbasis}.
At this point, the entire analysis would go through just as in the standard full-matrix Adagrad, 
except for the fact that the term $\left\langle g_{t},H_{t-1}^{-1}m_{t}\right\rangle $
may be very large whenever $H_{t-1}$ fails to align well with any
of $m_{t}$ and $g_{t}$. 

On the other hand, if we are willing to assume that these quantities are always well-aligned, in the sense that there exists an $\epsilon$ such that 
\[
\left\Vert \widetilde{g}_{t}\right\Vert _{H_{t-1}^{-1}}^{2}+\left\Vert T_{H_{t}}\left(g_{t}\right)\right\Vert _{H_{t-1}^{-1}}^{2}\leq\frac{k\epsilon}{2d}
\]
then we can obtain via a standard analysis with iterate bound $R_{\infty}$ that the regret $\mathcal{R}_{T}$ is bounded as 
\begin{align*}
\mathcal{R}_{T} & \leq \frac{1}{2}R_{\infty}^{2}\left(\delta d+\eta\sum_{t=0}^{T-1}\left(\left\Vert \widetilde{g}_{t}\right\Vert _{2}^{2}+\left\Vert T_{H_{t}}\left(g_{t}\right)\right\Vert _{2}^{2}\right)\right)+T\frac{\epsilon}{2}+\frac{d}{k}\frac{1}{2\eta}\ln\det\left(I+\frac{\eta}{\delta}\sum_{t=0}^{T-1}\left(\widetilde{g}_{t}\widetilde{g}_{t}^{\top}+T_{H_{t}}\left(g_{t}\right)T_{H_{t}}\left(g_{t}\right)^{\top}\right)\right)\\
 & \leq\widetilde{O}\left(R_{\infty}\sqrt{\sum_{t=0}^{T-1}\left(\left\Vert \widetilde{g}_{t}\right\Vert _{2}^{2}+\left\Vert T_{H_{t}}\left(g_{t}\right)\right\Vert _{2}^{2}\right)\cdot d\cdot\frac{d}{k}}+T\epsilon\right)
\end{align*}

Thus, provided that the truncated gradient norms are $O\left(1\right)$
this upper bound is 
\[
\epsilon+\widetilde{O}\left(\frac{R_{\infty}\sqrt{T\cdot d\cdot\frac{d}{k}}}{T }\right)
\]
 which follows the non-truncated version, after increasing the number
of iterations by a worst-case factor of $\frac{d}{k}$. 
In turn, this factor is natural, and also occurs in the worst-case analyses of error feedback without pre-conditioning, e.g.~\cite{2018-alistarh}.

\end{document}